\documentclass[a4paper,fleqn]{cas-dc}

\usepackage[numbers]{natbib}

\usepackage{amssymb}
\usepackage{graphicx,lipsum,wrapfig,subcaption}

\usepackage[nolist]{acronym}
\usepackage{url}
\usepackage{paralist}
\usepackage{hyperref}
\usepackage[T1]{fontenc}   
\usepackage{listings}

\usepackage{multirow}
\usepackage{colortbl}
\usepackage{hhline}

\newcommand{\quotes}[1]{``#1''}
\newtheorem{hypothesis}{Hypothesis}
\newtheorem{definition}{Definition}

\def\tsc#1{\csdef{#1}{\textsc{\lowercase{#1}}\xspace}}
\tsc{WGM}
\tsc{QE}

\begin{document}
	\let\WriteBookmarks\relax
	\def\floatpagepagefraction{1}
	\def\textpagefraction{.001}
	
	\shorttitle{An Objective Metric for Explanations and Explainable AI}    
	\shortauthors{F. Sovrano et al.}
	
	\title[mode = title]{An Objective Metric for Explainable AI:\\ How and Why to Estimate the Degree of Explainability}
	
	\author[disi]{Francesco Sovrano}[orcid=0000-0002-6285-1041]
	\ead{francesco.sovrano2@unibo.it}
	\cormark[1]
	
	\author[disi]{Fabio Vitali}[orcid=0000-0002-7562-5203]
	\ead{fabio.vitali@unibo.it}
	
	\affiliation[disi]{
		organization={DISI, University of Bologna},
		addressline={Mura Anteo Zamboni 7}, 
		city={Bologna},
		postcode={40126}, 
		country={Italy}
	}
	
	\cortext[1]{Corresponding author}
	
	\begin{abstract}
		Explainable AI was born as a pathway to allow humans to explore and understand the inner working of complex systems. 
		However, establishing what \textit{is} an explanation and \textit{objectively} evaluating \textit{explainability} are not trivial tasks. 
		This paper presents a new model-agnostic metric to measure the \acl{DoX} of information in an \textit{objective} way. We exploit a specific theoretical model from Ordinary Language Philosophy called the \textit{Achinstein's Theory of Explanations}, implemented with an algorithm relying on deep language models for knowledge graph extraction and information retrieval.
		To understand whether this metric can measure explainability, we devised a few experiments and user studies involving more than 190 participants, evaluating two realistic systems for \textit{healthcare} and \textit{finance} using famous AI technology, including Artificial Neural Networks and TreeSHAP.
		The results we obtained are statistically significant (with $P$ values lower than $.01$), suggesting that our proposed metric for measuring the \acl{DoX} is robust in several scenarios, and it aligns with concrete expectations. 
	\end{abstract}
	
	
	\begin{highlights}
		\item Presentation of a model-agnostic and deterministic metric for explainability: DoX.
		\item DoX is the first explainability metric based on Ordinary Language Philosophy.
		\item DoX can quantify Carnap's central criteria of explication adequacy.
		\item Presentation of an open-source software implementation of DoX called DoXpy.
		\item Evaluation of DoX with two user studies and more than 190 participants.
	\end{highlights}
	
	
	\maketitle
	
	\section{Introduction}\label{sec:introduction}
	Recent advances in Artificial Intelligence (AI) enable computer science and engineering to create machines that can learn from rough data, automating tasks previously thought to be accessible only by biological intelligence.
	However, these advances and results seem to come at a cost in terms of explainability, so the most effective machine learning techniques are, so far, not easily interpretable in symbolic terms \cite{chakraborty2017interpretability,rudin2019stop}. 
	
	The paradigms that address this explainability problem fall into the so-called \ac{XAI} field, which is broadly recognized as a crucial feature for the practical implementation of artificial intelligence models \cite{arrieta2020explainable}.
	Recently we are seeing a growing demand for explainability in AI applications, motivated by the growing realization that transparency is critical for fairness and legality.
	
	More precisely, in the European Union (EU), we now have several laws in force which establish obligations of explainability based on who uses AI (e.g., public authorities, private companies) and the degree of automation of the decision-making process (e.g., fully or partially automated) \cite{bibal2021legal}.
	As a result, the EU is indirectly posing an exciting challenge to the \acf{XAI} community by calling for more transparent, user-centered, and accountable \textit{automated decision-making systems} to ensure the explainability of their workings.
	
	In a recent attempt to capture the \quotes{legal requirements on explainability in machine learning}, \citeauthor{bibal2021legal} \cite{bibal2021legal} have identified four primary explainability requisites for Business-to-Consumer and Business-to-Business.
	In particular, \citeauthor{bibal2021legal} assert that, for Business-to-Consumer and Business-to-Business, explanations about a solely-automated decision-making system should at least provide information about:
	\begin{itemize}
		\item the main features used in a decision taken by the AI;
		\item all features processed by the AI;
		\item the specific decision taken by the AI;
		\item the underlying logical model followed by the AI.
	\end{itemize}
	Therefore, with the present paper, we want to expand further the work of \citeauthor{bibal2021legal}, trying to understand whether it is possible to objectively quantify how much of the information required by the law is explained by an AI.
	
	In this paper, we propose a new model-agnostic approach and metric to \textit{objectively} evaluate explainability in a manner mainly inspired by Ordinary Language Philosophy instead of Cognitive Science.
	Our approach is based on a specific theoretical model of explanation, called the \textit{Achinstein's theory of explanations}, where explanations are the result of an \textit{illocutionary} (i.e., broad yet pertinent and deliberate) act of pragmatically answering to a question. 
	Accordingly, explanations are answers to many basic questions (\textit{archetypes}), each of which sheds a different light on the concepts being explained. 
	As a consequence, the more (archetypal) answers an \textit{automated decision-making system} can give about the important aspects of its explanandum\footnote{The word \textit{explanandum} means \quotes{what is to be explained}, in Latin.}, the more it is explainable. 
	
	Therefore, we assert that it is possible to quantify the degree of explainability of a set of texts by applying the Achinstein-based definition of explanation proposed in \cite{sovrano2022generating}.
	Thus, drawing also from Carnap's criteria of adequacy of an explication \cite{novaes2017carnapian}, we frame the \ac{DoX} as the average \textit{explanatory illocution} of information on a set of \textit{explanandum aspects}\footnote{Carnap uses the term \textit{explicandum} where we employ \textit{explanandum}, but, by and large, we assume the two words can be used interchangeably. They both mean \quotes{what needs to be explained} in Latin.}.
	More precisely, we hereby present an algorithm for measuring explainability through pre-trained \textit{deep language models} for general-purpose answer retrieval (e.g., \cite{karpukhin2020dense,bowman2015large}) applied to a particular graph of triplets automatically extracted from text to facilitate this type of information retrieval.
	
	Hence, we made the following hypothesis.
	\begin{hypothesis} \label{hyp:main}
		\textbf{\ac{DoX} scores measure explainability}: a \ac{DoX} score can describe \textit{explainability}, so that, given the same explanandum, a higher \ac{DoX} implies greater \textit{explainability} and a lower \ac{DoX} implies smaller \textit{explainability}.
	\end{hypothesis}
	
	To verify this hypothesis, we devised and implemented a pipeline of algorithms called DoXpy to compute \textit{DoX} scores.
	We also performed a few experiments to show that \textit{explainability} changes in accordance with varying \ac{DoX} scores.
	Notably, the results of all our experiments clearly and undoubtedly showed that Hypothesis \ref{hyp:main} holds.
	
	This paper is structured as follows.
	In Section \ref{sec:background} we give the necessary background information to introduce the theoretical models properly discussed subsequently (i.e., Achinstein's theory), while in Section \ref{sec:related_work}, we discuss existing literature, comparing it to our proposed solution.
	In Section \ref{sec:proposed_solution}, we show how a metric for quantifying the degree of explainability is possible by defining \textit{explaining} as an illocutionary act of question-answering and by verifying Carnap's criteria employing deep language models.
	In Section \ref{sec:experiments}, we describe our experiments, discussing the results and some possible limitations of \ac{DoX} in Section \ref{sec:results_discussion_n_limitations}.
	Finally, we point to future work and conclusions in Section \ref{sec:conclusions}.
	
	To guarantee the reproducibility of the experiments, we publish the source code\footnote{\url{https://github.com/Francesco-Sovrano/DoXpy}} of the algorithm for computing \ac{DoX} scores, as well as the code of the systems used for the experiment, the full details of our user studies and the complete set of data mentioned in this paper.
	
	\section{Background} \label{sec:background}
	
	This section provides some background to justify and support the rest of the paper.
	Hereby we briefly summarise several recent and less recent approaches to the theories of explanation, with a particular focus on Achinstein's. After that, we discuss how Achinstein's theory of explaining as a question-answering process is compatible with existing \ac{XAI} literature, highlighting how profound the connection between answering questions and explaining is in this field.
	
	\subsection{Adequacy of Explainability: Carnap's Criteria} \label{sec:carnap}
	
	In philosophy, the most important work about the criteria of adequacy of \textit{explainable information} is likely to be Carnap's \cite{leitgeb2021carnap}. 
	Even though Carnap studies the concept of \textit{explication} rather than that of \textit{explainable information}, we assert that they share a common ground making his criteria fitting in both cases.
	\textit{Explication} in Carnap's sense is the replacement of a somewhat unclear and inexact concept, the \textit{explicandum}, by a new, clearer, and more exact concept, the \textit{explicatum}\footnote{i.e., \quotes{what has been explained}, in Latin.}, and this is precisely what information does when made explainable.
	
	Carnap's main criteria of explication adequacy\cite{leitgeb2021carnap} are \textit{similarity}, \textit{exactness} and \textit{fruitfulness}\footnote{Carnap also discussed another desideratum, \textit{simplicity}. However, this criterion is presented as subordinate to the others (especially exactness).}.
	\textit{Similarity} means that the explicatum should be \textit{detailed} about the explicandum, in the sense that at least many of the intended uses of the explicandum, brought out in the clarification step, are preserved in the explicatum.
	On the other hand, \textit{exactness} means that the explication should be embedded in some sufficiently \textit{clear} and exact linguistic framework, while \textit{fruitfulness} implies that the explicatum should be \textit{useful} and usable in a variety of other \textit{good} explanations (the more, the better). 
	
	Carnap's adequacy criteria are transversal to all the identified definitions of explainability, possessing preliminary characteristics for any information to be adequately considered explainable.
	Interestingly, the property of \textit{truthfulness} (being different from \textit{exactness}) is not explicitly mentioned in Carnap's desiderata.
	That is to say that explainability and \textit{truthfulness} are complementary but different, as also discussed by \cite{hilton1996mental}. An explanation is such regardless of its truth (high-quality but ultimately false explanations exist, especially in science). Vice versa, highly correct information can be inferior at explaining.
	
	\subsection{Definitions of Explainability} \label{sec:contemporary_theories_of_explanation}
	
	Considering the definition of \quotes{explainability} as \quotes{the potential of information to be used for explaining}, we envisage that a proper understanding of how to measure explainability must pass through a thorough definition of what constitutes an explanation and of the act of explaining.
	
	In 1948 Hempel and Oppenheim published their \quotes{Studies in the Logic of Explanation} \cite{hempel1948studies}, giving birth to what is considered the first theory of explanations: the deductive-nomological model.
	After that work, many amended, extended, or replaced this model, which came to be considered fatally flawed \cite{bromberger1966questions,salmon1984scientific}. Several more modern and competing theories of explanations resulted from this criticism.
	
	\begin{table*}[!htb]
		\caption{\textbf{Philosophical definitions of explanation and explainable information.} In this table, we summarise the definitions of \textit{explanation} and \textit{explainable information} for each one of the identified theories of explanations.} \label{tab:definitions}
		\makebox[\linewidth]{
			\begin{tabular}{|p{0.20\linewidth}|p{0.50\linewidth}|p{0.3\linewidth}|}
				\hline
				\rowcolor[HTML]{C0C0C0} 
				\textbf{Theory}              & \textbf{Explanations}                                                                                                                                      & \textbf{Explainable Information}                             \\ \hline
				Causal Realism \cite{salmon1984scientific}          & Descriptions of causality, expressed as chains of causes and effects.                                                                                                              & What can fully describe causality.                            \\ \hline
				Constructive Empiricism \cite{van1980scientific}     & Contrastive information that answers \texttt{why} questions, allowing one to calculate the probability of a particular event relative to a set of (possibly subjective) background assumptions.                                                   & What provides answers to contrastive \texttt{why} questions.                    \\ \hline
				Ordinary Language \linebreak Philosophy \cite{achinstein1983nature}   & Answers to questions (not just \texttt{why} ones) given with the explicit intent of producing understanding in someone, i.e., the result of an illocutionary act.                                                                                           & What can be used to pertinently answer questions about relevant aspects with \textit{illocutionary force}. \\ \hline
				Cognitive Science \cite{holland1986induction}        & Mental representations resulting from a cognitive activity. They are information which fixes failures in someone's mental model.                                              & What can have a \textit{perlocutionary effect}, fixing failures in someone's mental model.                          \\ \hline
				Naturalism and Scientific Realism \cite{sellars1963philosophy} & Information which increases the coherence of someone's belief system, resulting from an iterative process of confirmation of truths aimed at improving understanding. & What can have a \textit{perlocutionary effect}, increasing coherence of someone's belief system.  \\ \hline
			\end{tabular}
		}
	\end{table*}
	
	Summarising our full analysis \cite{sovrano2022survey}, the five most important theories of explanation in contemporary philosophy are:
	\begin{inparadesc}
		\item Causal Realism,
		\item Constructive Empiricism, 
		\item Ordinary Language Philosophy,
		\item Cognitive Science,
		\item Naturalism and Scientific Realism.
	\end{inparadesc}
	Consequently, there are at least five definitions of \quotes{explanation}, one per theory.
	A summary of these definitions is shown in Table \ref{tab:definitions}, highlighting that there is no complete agreement between them on the nature of explanations.
	
	In particular, Hempel's, Salmon's (Causal Realism), and Van Fraassen's (Constructive Empiricism) theories frame the act of explaining more as a \textit{locutionary act} \cite{austin1975things}, whereby an explanation is such because it utters something.
	Differently, Achinstein's theory (from Ordinary Language Philosophy) explicitly frames explaining as an \textit{illocutionary act} \cite{austin1975things} so that an explanation is such because of the intention to explain.
	The theories of Holland (Cognitive Science) and Sellars (Naturalism/Scientific Realism), on the other hand, frame explaining more as a \textit{perlocutionary act} \cite{austin1975things}, thus with an explanation being such because of the effects it produces in the interlocutor.
	
	Notably, we notice that whenever explaining is considered to be an act that has to satisfy someone's needs, then explainability differs from explaining.
	In fact, in this context, pragmatically satisfying someone (i.e., user-centrality) is achieved when explanations are tailored to a specific person so that the same explainable information can be presented and re-elaborated differently across different individuals.
	It follows that in each philosophical tradition except Salmon's Causal Realism \cite{salmon1984scientific}, we have a definition of \quotes{explainable information} that slightly differs from that of \quotes{explanation}, as described in \cite{sovrano2022survey}. 
	For example, in Ordinary Language Philosophy \textit{explainable information} can be understood as \quotes{what can be used to pertinently answer questions about relevant aspects, in an illocutionary way}.
	
	\subsection{Explainability According to Ordinary Language Philosophy} \label{sec:archetypal_questions}
	
	According to Achinstein's theory, explanations result from an \textit{illocutionary} act of pragmatically answering a question. 
	In particular, it means that there is a subtle and essential difference between simply \quotes{answering to questions} and \quotes{explaining}, and this difference is \textit{illocution}.
	
	It appears that an \textit{illocutionary} act results from a clear intent of achieving the goal of such act, as a promise being \quotes{what it is} just because of the intent of maintaining it. So that \textit{illocution} in explaining makes an explanation as such just because it is the result of an underlying and proper intent of explaining.
	
	Despite this definition, \textit{illocution} seems too abstract to implement inside an actual software application.
	Nonetheless, recent efforts towards the automated generation of explanations \cite{sovrano2021philosophy,sovrano2022generating}, have shown that it may be possible to define \textit{illocution} in a more \quotes{computer-friendly} way.
	Indeed, as stated in \cite{sovrano2022generating}, illocution in explaining involves informed and \textit{pertinent} answers not just to the main question but also to other questions of various kinds, even unrelated to causality that are relevant to the explanations.
	These questions can be understood as instances of archetypes such as \texttt{why}, \texttt{why not}, \texttt{how}, \texttt{what for}, \texttt{what if}, \texttt{what}, \texttt{who}, \texttt{when}, \texttt{where}, \texttt{how much}, etc.
	
	\begin{definition}[Archetypal Question] \label{def:archetypal_question}
		An \textit{archetypal question} is an archetype applied to a specific aspect of the explanandum. 
		Examples of archetypes are the interrogative particles (e.g., \texttt{why}, \texttt{how}, \texttt{what}, \texttt{who}, \texttt{when}, \texttt{where}), or their derivatives (e.g., \texttt{why not}, \texttt{what for}, \texttt{what if}, \texttt{how much}), or also more complex interrogative formulas (e.g., \texttt{what reason}, \texttt{what cause}, \texttt{what effect}).
		Accordingly, the same archetypal question may be rewritten in several different ways, as \quotes{why} can be rewritten in \quotes{what is the reason} or \quotes{what is the cause}.
	\end{definition}
	
	Thus, archetypal questions provide generic explanations on a specific aspect of the explanandum in a given informative context, which can precisely link the content to the informative goal of the person asking the question.	
	For example, if the explanandum were \quotes{heart diseases}, there would be many aspects involved, including \quotes{heart}, \quotes{stroke}, \quotes{vessels}, \quotes{diseases}, \quotes{angina}, \quotes{symptoms}. Some archetypal questions, in this case, are \quotes{What is angina?} or \quotes{Why a stroke?}.
	
	
	\subsection{Explainable AI and Question Answering} \label{sec:xai_as_qa}
	
	Suppose we assume that the interpretation of Achinstein's theory of explanations given by \cite{sovrano2022generating} is correct. In that case, data or processes are said to be \textit{explainable} when their informative content can adequately answer \textit{archetypal questions}. 
	
	The idea of answering questions as explaining is not new to the field of \ac{XAI} \cite{liao2020questioning}, and it is also compatible with our intuition of what constitutes an explanation.
	It is common to many works in the field \cite{ribera2019can,lim2009and,miller2018explanation,gilpin2018explaining,dhurandhar2018explanations, wachter2017counterfactual,rebanal2021xalgo,jansen2016s,madumal2019grounded} the use of generic (e.g., \texttt{why}, \texttt{who}, \texttt{how}, \texttt{when}) or more punctual questions to clearly define and describe the characteristics of explainability \cite{liao2020questioning}.
	
	For example, \citeauthor{lundberg2020local} \cite{lundberg2020local} assert that the local explanations produced by their TreeSHAP (an \textit{additive feature attribution} method for feature importance) may \quotes{help human experts understand \textit{why} the model made a specific recommendation for high-risk decisions}.
	On the other hand, \citeauthor{dhurandhar2018explanations} \cite{dhurandhar2018explanations} clearly state that they designed CEM (a method for the generation of counterfactuals and other contrastive explanations) to answer the question \quotes{why is input \textit{x} classified in class \textit{y}?}.
	Furthermore, \citeauthor{rebanal2021xalgo} \cite{rebanal2021xalgo} propose and studies an interactive approach where explaining is defined in terms of answering \texttt{why}, \texttt{what} and \texttt{how} questions.
	These are just some examples, among many, of how Achinstein's theory of explanations is already implicit in existing \ac{XAI} literature. They highlight how deep the connection between answering questions and explaining is in this field.
	
	Nonetheless, despite the compatibility, practically none of the works in \ac{XAI} explicitly mentions any theory from Ordinary Language Philosophy, preferring to refer to Cognitive Science \cite{miller2018explanation,hoffman2018metrics} instead.
	This is probably because Achinstein's illocutionary theory of explanations is challenging to implement into software by being utterly pragmatic. 
	\emph{User-orientedness} is challenging and sometimes not connected to the primary goal of \ac{XAI}: \quotes{opening the black box} (e.g., understanding how and why an opaque AI model works).
	
	\section{Related Work} \label{sec:related_work}
	
	\begin{table*}[!htb]
		\centering
		\caption[Comparison of Different Explainability Metrics]{
			\textbf{Comparison of Different Explainability Metrics}\footnotemark: The column \quotes{Sources} points to referenced papers, while column \quotes{Metrics} points to the names of the metrics.
			Elements in bold are column-by-column better than the rest.
		}
		\label{tab:literature_comparison}
		\makebox[\linewidth]{
			\begin{tabular}{|p{.11\linewidth}|p{.13\linewidth}|p{.2\linewidth}|p{.07\linewidth}|p{.12\linewidth}|p{.25\linewidth}|}
				\hline
				\rowcolor[rgb]{0.753,0.753,0.753} \textbf{Source}                                                                                                                          & \textbf{Model \& Information Format}          & \textbf{Closest Supporting Theory}                                                                          & \textbf{Subject - based} & \textbf{Measured Carnap's Criteria}                                                                               & \textbf{Metrics}                                                                                                       \\ 
				\hline
				\cite{rosenfeld2021better}                                                                                                                                & Rule-based                           & Causal Realism                                                                                      & \textbf{No}              & \begin{tabular}[c]{@{}l@{}}Exactness,\\Fruitfulness\end{tabular}                                           & \begin{tabular}[c]{@{}l@{}}Performance Difference, \\Number of Rules,\\Number of Features, Stability\end{tabular}  \\ 
				\hline
				\cite{villone2020comparative}                                                                                                                             & Rule-based                           & Causal Realism                                                                                      & \textbf{No}              & \begin{tabular}[c]{@{}l@{}}Similarity,\\Fruitfulness\end{tabular}                                        & \begin{tabular}[c]{@{}l@{}}Fidelity, \\ Completeness\end{tabular}                                                   \\ 
				\hline
				\cite{nguyen2020quantitative}                                                                                                                             & Feature Attribution                  & Causal Realism                                                                                      & \textbf{No}              & \begin{tabular}[c]{@{}l@{}}Exactness,\\Fruitfulness\end{tabular}                                        & \begin{tabular}[c]{@{}l@{}}Monotonicity, Non-sensitivity, \\ Effective Complexity\end{tabular}                   \\ 
				\hline
				\cite{lakkaraju2017interpretable}                                                                                                                         & Rule-based                           & Causal Realism                                                                                      & \textbf{No}              & \begin{tabular}[c]{@{}l@{}}\textbf{Similarity,}\\\textbf{Exactness,}\\\textbf{Fruitfulness}\end{tabular} & \begin{tabular}[c]{@{}l@{}}Fidelity, Unambiguity, \\ Interpretability, Interactivity\end{tabular}                \\ 
				\hline
				\cite{holzinger2020measuring}                                                                                                                             & \textbf{Any}                         & \begin{tabular}[c]{@{}l@{}}Causal Realism, \\ Cognitive Science, \\ Naturalism \& Co.\end{tabular} & Yes                      & \begin{tabular}[c]{@{}l@{}}Exactness,\\Fruitfulness\end{tabular}                                         & System Causability Scale                                                                                                         \\ 
				\hline
				\cite{hoffman2018metrics}                                                                                                                                 & \textbf{Any}                         & \begin{tabular}[c]{@{}l@{}}Cognitive Science, \\ Naturalism \& Co.\end{tabular}                    & Yes                      & \begin{tabular}[c]{@{}l@{}}Exactness,\\Fruitfulness\end{tabular}                                         & \begin{tabular}[c]{@{}l@{}}Satisfaction, Trust, \\ Mental Models, \\ Curiosity, Performance\end{tabular}            \\ 
				\hline \cite{dieber2022novel,sovrano2021philosophy,mohseni2021quantitative,wang2021explanations,szymanski2021visual,buccinca2020proxy,poursabzi2018manipulating} & \textbf{Any}                         & \begin{tabular}[c]{@{}l@{}}Cognitive Science, \\ Naturalism \& Co.\end{tabular} & Yes                      & \begin{tabular}[c]{@{}l@{}}Exactness,\\Fruitfulness\end{tabular} & \begin{tabular}[c]{@{}l@{}}Usability: Effectiveness, \\Efficiency, Satisfaction\end{tabular}                    \\ 
				\hline
				\cite{arras2022clevr}                                                                                                                             & Heatmap                  & Constructive Empiricism                                                                                      & \textbf{No}              & \begin{tabular}[c]{@{}l@{}}Similarity, \\Exactness\end{tabular}                                        & \begin{tabular}[c]{@{}l@{}}Relevance Mass Accuracy, \\Relevance Rank Accuracy\end{tabular}                   \\ 
				\hline
				\cite{keane2021if}                                                                                                                                        & Prototype-based                        & Constructive Empiricism                                                                             & \textbf{No}              & Exactness                                                                                                & \begin{tabular}[c]{@{}l@{}}Proximity, Sparsity, \\ Adequacy (Coverage)\end{tabular}                                 \\ 
				\hline
				\cite{nguyen2020quantitative}                                                                                                                             & Prototype-based                        & Constructive Empiricism                                                                             & \textbf{No}              & \begin{tabular}[c]{@{}l@{}}Similarity,\\Fruitfulness\end{tabular}                                        & \begin{tabular}[c]{@{}l@{}}Non-Representativeness, \\ Diversity\end{tabular}                                        \\ 
				\hline
				This Paper                                                                                                                                  & \textbf{Any} (Natural Language Text) & \begin{tabular}[c]{@{}l@{}}Ordinary Language \\Philosophy\end{tabular}                                                                                  & \textbf{No}              & \begin{tabular}[c]{@{}l@{}}\textbf{Similarity,}\\\textbf{Exactness,}\\\textbf{Fruitfulness}\end{tabular} & Degree of Explainability                               \\
				\hline
			\end{tabular}
		}
	\end{table*}
	\footnotetext{This table extends a similar one in \cite{sovrano2022survey}.}
	
	Measuring the quality of explanations and \ac{XAI} tools is pivotal for claiming technological advancements, understanding existing limitations, developing better solutions, and delivering \ac{XAI} that can go into production.
	Not surprisingly, every good paper proposing a new \ac{XAI} algorithm comes with evidence and experiments backing up their claims and none other, usually relying on \textit{ad hoc} or subjective mechanisms for measuring the quality of their explainability. This makes it very hard to perform meaningful comparisons. 
	
	In other words, as also suggested by literature reviews (e.g.,  \cite{vilone2021notions}, and especially \cite{sovrano2022survey}, which reports in Table \ref{tab:literature_comparison} its main results), it is common to encounter explainability metrics that work only with a specific \ac{XAI} model or prove their usefulness by collecting human-generated opinions/results after interacting with the studied system and no other. 
	
	For example, the metrics proposed in \cite{arras2022clevr,rosenfeld2021better,villone2020comparative,nguyen2020quantitative,lakkaraju2017interpretable,keane2021if} can only be used with specific types of \ac{XAI} approaches (e.g., prototype selection or feature attribution). Instead, the metrics proposed in \cite{hoffman2018metrics,holzinger2020measuring,dieber2022novel} rely on user studies, as many other works \cite{sovrano2021philosophy,mohseni2021quantitative,wang2021explanations,szymanski2021visual,buccinca2020proxy,poursabzi2018manipulating}, based on classical usability metrics (i.e., effectiveness, efficiency, satisfaction).
	
	Only one work among those examined, \cite{hoffman2018metrics}, claims its proposed metric is model-agnostic and thus generic enough to be compatible with any \ac{XAI}.
	In particular, this is possible because the work measures explainability \textit{indirectly} by estimating the effects of explanations on human subjects.
	More precisely, \cite{hoffman2018metrics} is mainly inspired by the interpretation of explanations given by Cognitive Science, requiring measuring:
	\begin{inparaenum}[i)]
		\item the subjective goodness of explanations;
		\item whether users are satisfied by explanations;
		\item how well users understand the AI systems;
		\item how curiosity motivates the search for explanations;
		\item whether the user's trust and reliance on the AI are appropriate;
		\item how the human-\ac{XAI} work system performs.
	\end{inparaenum}
	
	Indeed, the metric presented in \cite{hoffman2018metrics} is non-deterministic and heavily relies on subjective measurements, despite being model-agnostic. 
	The metric we propose here, \ac{DoX}, is objective, deterministic, and model-agnostic\footnote{\ac{DoX} is model-agnostic only under the assumption that any explanation or bit of explainable information can be represented or described in natural language, e.g., English.}. It can be used to evaluate the explainability of any textual information and to understand whether the amount of explainability is objectively poor, even if the explanations are perceived as satisfactory and sound by the explainees.
	
	Furthermore, only \ac{DoX} and \cite{lakkaraju2017interpretable} appear to measure all three main Carnap's desiderata.
	More specifically, \citeauthor{lakkaraju2017interpretable} \cite{lakkaraju2017interpretable} evaluate Carnap's criteria separately, while with \ac{DoX}, we propose a single metric that combines all of them.
	
	Finally, as suggested in \cite{sovrano2022survey}, all existing explainability metrics can be aligned to different interpretations of explainability coming from complementary theories of explanations.
	As shown in Table \ref{tab:literature_comparison}, most of these metrics seem aligned with Causal Realism and Cognitive Science. In contrast, \ac{DoX} is the first metric based on Ordinary Language Philosophy.
	
	
	\section{Degrees of Explanation (DoX)} \label{sec:proposed_solution}
	
	In Section \ref{sec:related_work}, we discussed how existing metrics for measuring (properties of) explainability are frequently either model-specific or subjective, raising the question of whether it is possible to measure the degree of explainability with fully automated software objectively.
	With this paper, we try to answer this question by leveraging on an extension of Achinstein's theory of explanations as proposed in \cite{sovrano2022generating} and summarized in Section \ref{sec:archetypal_questions}.
	We do it by asserting that any algorithm for measuring the degree of explainability must pass through a thorough definition of what constitutes \textit{explainability} and \textit{explanation}.
	Considering that \textit{explainability} is fundamentally the \textit{ability to explain}, it is clear that a proper definition of it requires a precise understanding of what is \textit{explaining}.
	
	In this section, we discuss the theory behind \ac{DoX} and a concrete implementation to measure \ac{DoX} in practice.
	
	\subsection{Quantifying the Degree of Explainability} \label{sec:proposed_solution:theory}
	
	As discussed in Section \ref{sec:xai_as_qa}, the informative contents of state-of-the-art \ac{XAI} are clearly polarised towards answering \texttt{why}, \texttt{what if} or \texttt{how} questions. 
	Considering that \texttt{why}, \texttt{what if}, and \texttt{how} are different questions pointing to different types of information, which type is the best one? 
	We assert that the correct answer to this question is: \quotes{none}. 
	Depending on the needs of the explainees, their background knowledge, the context, and potentially many other factors, each archetype may be equally important.
	
	In other words, depending on the characteristics of the explainee (e.g., background knowledge, objectives, context), a combination of different \ac{XAI} mechanisms may be necessary to obtain a minimum \textit{understanding of the internal logic of a black-box AI}. 
	Therefore, knowing the types of explainability covered by a system using XAI can be of the utmost importance in understanding how explainable it is.
	Hence, following this intuition, we started to study how to measure explainability in terms of (generic) questions.
	
	Among the different approaches mentioned in Section \ref{sec:contemporary_theories_of_explanation}, the closest one to our intuition of explainability is probably Achinstein's theory, coming from Ordinary Language Philosophy.
	Achinstein defines the act of explaining as an act of illocutionary question-answering, stating that \textit{explaining} is more than \textit{answering a question} because it requires some form of illocution.
	Nonetheless, without a precise and computer-friendly definition of illocution, it is hard to go further than a philosophical and abstract understanding of such a concept.
	For this reason, as discussed in Section \ref{sec:archetypal_questions}, \cite{sovrano2021philosophy} suggested that illocution (or, better, \textit{explanatory illocution}) is, in fact, the process of answering multiple generic and primitive questions (e.g., \texttt{why}, \texttt{how}, \texttt{what}) called \textit{archetypal questions}.
	
	For example, if someone is asking \quotes{How are you doing?}, an answer like \quotes{I am good} would not be considered an explanation.
	Differently, the answer \quotes{I am happy because I just got a paper accepted at this important venue, and [...]} would instead be normally considered an explanation because it answers other \textit{archetypal questions} together with the main question.
	
	We are convinced that, under these premises, we can concretely measure the degree of explainability of information quantitatively.
	More precisely, we propose that the degree of explainability of the information depends on the number of \textit{archetypal questions} to which it can adequately answer.
	In other words, we estimate the degree of explainability of a piece of information by measuring its relevance to answering a (pre-defined) set of archetypal questions.
	
	Therefore, our theoretical contribution, set out in the following subsections, consists of the precise and formal definition of:
	\begin{inparadesc}
		\item \textit{cumulative pertinence},
		\item \textit{explanatory illocution},
		\item \textit{\acf{DoX}},
		\item and \textit{average \ac{DoX}}.
	\end{inparadesc}
	We will first provide formal definitions and then explain them further with some examples.
	
	\subsubsection{Cumulative Pertinence, Explanatory Illocution and DoX}
	
	Assuming the correctness of a given piece of information, explainability is a property of that information. 
	Explainability can be measured in terms of \textit{explanatory illocution}.
	In order to understand what explanatory illocution is, we have to define the concept of \textit{cumulative pertinence} first.
	
	\begin{definition}[Cumulative Pertinence]
		The \textit{cumulative pertinence} is an estimate of how \textit{pertinently} and how in \textit{detail} a given piece of information $\Phi$ can answer a question about an aspect $a$ of an explanandum $\Delta$. 
		Let $A$ be the set of relevant aspects to be explained about $\Delta$.
		Let $D_a$ be the subset of all the details (e.g., sentences, grammatical clauses\footnote{A typical \textit{clause} consists of a subject and a syntactic predicate, the latter typically a verb phrase composed of a verb with any objects and other modifiers.}, paragraphs) in $\Phi$ that are about an aspect $a \in A$.
		Let $q_a$ be a question about an aspect $a \in A$.
		Let $p\left(d,q_a\right) \in [0,1]$ be the pertinence of a detail $d \in D_a$ to $q_a$. 
		Let also $t$ be a pertinence threshold in the $[0,1]$ range.
		Then, the \textit{cumulative pertinence} of $D_a$ to $q_a$ is $P_{D_a,q_a} = \sum_{d \in D_a, p\left(d,q_a\right) \geq t}{p\left(d,q_a\right)}$. 
	\end{definition}
	
	\begin{definition}[Explanatory Illocution - Formal Definition] \label{def:explanatory_illocution}
		The \textit{explanatory illocution} is a set of \textit{cumulative pertinences} for a pre-defined set of \textit{archetypal questions}. 
		Let $Q$ be a set of archetypes $q$ and $q_a$ be the question obtained by applying the archetype $q$ to an aspect $a \in A$.
		Then the \textit{explanatory illocution} of $\Phi$ to an aspect $a \in A$ is the set of tuples $\{\forall q \in Q \vert <q, P_{D_a,q_a}> \}$\footnote{The operator $<x, y>$ is used here to represent tuples.}. 
	\end{definition}
	
	
	Consequently, we define \ac{DoX} as follows.
	
	\begin{definition}[Degree of Explainability] \label{def:DoX}
		\ac{DoX} is the average \textit{explanatory illocution} per archetype, on the whole set $A$ of relevant aspects to be explained.
		In other terms, let $R_{D,q,A} = \frac{\sum_{a \in A}{P_{D_a,q_a}}}{|A|}$ be the average \textit{cumulative pertinence} of $D$ to $q$ and $A$, where $D = \{\forall a \in A, \forall d \in D_a \vert d \}$, then the \ac{DoX} is the set $\{\forall q \in Q \vert <q, R_{D,q,A}> \}$.
	\end{definition}
	
	However, \ac{DoX} alone cannot help in judging whether some collections of information have higher degrees of explainability than others.
	This is because \ac{DoX} is a set, and sets are not sortable. 
	Thus we combine the set of pertinence scores composing \ac{DoX} into a single score representing explainability, called \textit{average \ac{DoX}}.
	So, the resulting \textit{average \ac{DoX}} can act as a metric to judge whether the explainability of a system is greater than, equal to, or lower than another.
	
	\begin{definition}[Average Degree of Explainability] \label{def:AveDoX}
		The Average \ac{DoX} is the average of the pertinence of each archetype composing the \ac{DoX}. 
		In other terms, the Average \ac{DoX} is $\frac{\sum_{q \in Q}{R_{D,q,A}}}{|Q|}$.
	\end{definition}
	
	The average \ac{DoX} represents a naive approach to quantify explainability with a single score, as it implies that all the archetypal questions and aspects have the same weight. However, this may not necessarily be true.
	As suggested by \citeauthor{liao2020questioning} \cite{liao2020questioning}, it seems that there is a shared understanding that \texttt{why} explanations are the most important in \ac{XAI}, sometimes followed by \texttt{how}, \texttt{what for}, \texttt{what if} and, possibly, \texttt{what}. 
	In other words, the relevance of an explanation can be estimated by the ability to effectively answer the most relevant (archetypal) questions for the stakeholders' objectives.
	Nonetheless, defining which (archetypal) question is the most relevant is challenging and somewhat subjective. Therefore we believe that average \ac{DoX} is probably the only objective solution to this dispute.
	
	%
	
	We will now discuss some examples of applying the formulas mentioned earlier. We will also demonstrate how these formulas can measure Carnap's adequacy criteria.
	
	\subsubsection{Interpreting DoX in Terms of Carnap's Criteria} \label{sec:theory:dox:formula:carnap}
	
	Suppose the sentence \quotes{I am happy that my article has been accepted in this prestigious journal} is given as $\Phi$ and the set of relevant aspects \textit{\{heart, stroke, vessel, disease, angina, symptom\}} as $A$.
	In this case, the set of details $D$ contains the following details:
	\begin{itemize}
		\item \quotes{I am happy};
		\item \quotes{my article has been accepted in this prestigious journal};
		\item \quotes{I am happy that my article has been accepted}.
	\end{itemize}
	However, none of the details above is about the explanandum. Thus $D_a = \emptyset, \forall a \in A$, because nothing in $\Phi$ is related to $A$. 
	Hence, the average cumulative pertinence would be equal to $0$ for every archetype $q \in Q$, forcing the \ac{DoX} score to be equal to $0$, as expected.
	In other words, no detail of $\Phi$ would explain anything about $A$. Therefore the explainability of $\Phi$ for $A$ would be zero.
	
	On the contrary, we would not have a null \ac{DoX} for $A$ when using the sentence \quotes{angina happens when some part of your heart does not get enough oxygen} as $\Phi$.
	That is because the new $\Phi$ contains details about at least two relevant aspects in $A$: \quotes{angina}, \quotes{heart}.
	Such details would score a higher average cumulative pertinence $R_{D,q,A}$ for $q$ equal to \texttt{why} because they are about causality.
	
	Eventually, when computing the \ac{DoX} of the new $\Phi$ for this set of explanandum aspects $A$ with the DoXpy algorithm presented in Section \ref{sec:proposed_solution:practice}\footnote{When using the MiniLM pertinence estimator introduced in Section \ref{sec:proposed_solution:practice:pertinence_functions}.}, the \textit{average \ac{DoX}} is $0.29$.
	In particular, as expected, the archetypes with the best score are the ones related to causality (i.e., \texttt{what effect} has a score of $0.59$; \texttt{in what case}, \texttt{why} and \texttt{how} have a score of $0.57$). In contrast, most of the other archetypes have a null score (i.e., \texttt{who}, \texttt{when}).
	
	Given Definition \ref{def:DoX}, we can say that \ac{DoX} is an estimate of the \textit{fruitfulness} of $D$ that combines in one single score the \textit{similarity} of $D$ to $A$ and the \textit{exactness} of $D$ for $Q$.
	For these reasons, \ac{DoX} is akin to Carnap's \textit{central} criteria of adequacy of explanation (introduced in Section \ref{sec:carnap}).
	Although, differently from Carnap, our understanding of \textit{exactness} is not that of adherence to standards of formal concept formation\footnote{Actually, Carnap did not specify what he means by \quotes{exactness}. Regardless, in this context, \quotes{exactness} is often viewed as either lack of vagueness or adherence to standards of formal concept formation.} \cite{brun2016explication}, but rather that of being precise or pertinent enough as an answer to a given question.
	
	The number of relevant explanandum aspects covered by a given piece of information, and the number of details that are pertinent about it (i.e., $\vert \{\forall a \in A, \forall d \in D_a \vert d \} \vert$), roughly say how much \textit{similar} that information is to the explanandum.
	More precisely, the formula used for computing the cumulative pertinence $P_{D_a,q_a}$ sums the contribution of every single detail according to its pertinence to the aspects $a \in A$, telling us how much $D_a$ is similar to $a$. 
	Thus, if \textit{pertinence} $p\left(d,q_a\right)$ would close to zero for all archetypes $q \in Q$, then a detail $d$ would have nothing to do with an aspect $a$.
	Furthermore, the average cumulative pertinence $R_{D,q,A}$ contains information about the \textit{exactness} of multiple answers, aggregating pertinence scores. 
	As a result, by measuring $R_{D,q,A}$ for all the $q \in Q$, we also obtain an estimate of how $D$ is \textit{fruitful} for the formulation of many other different explanations intended as the result of an illocutionary act of pragmatically answering questions.
	
	This construction of \ac{DoX} in terms of Carnap's main criteria (cf. Section \ref{sec:carnap}) of adequacy is crucial because it allows implementing an actual algorithm to quantify explainability as shown by the experimental results presented in Section \ref{sec:experiments}.
	
	
	
	\subsection{The DoXpy Algorithm} \label{sec:proposed_solution:practice}
	
	Throughout this section, we will explain why and how to use existing algorithms for answer retrieval and information extraction to implement DoXpy, an algorithm for computing \ac{DoX}.
	We publish the DoXpy source code at \url{https://github.com/Francesco-Sovrano/DoXpy} for reproducibility purposes.
	
	Given Definition \ref{def:DoX}, we argue that it is possible to write an algorithm that can approximately quantify the \acl{DoX} of information representable with \textit{natural language} (e.g., English) by adapting existing technology for question-answering.
	In particular, according to Definition \ref{def:explanatory_illocution}, in order to implement an algorithm capable of computing the (average) \ac{DoX} of $\Phi$, we need to:
	\begin{itemize}
		\item define a set $A$ of \textit{explanandum aspects};
		\item identify the set of all possible archetypes $Q$;
		\item define a mechanism to identify the set $D$ of details contained in $\Phi$ and the subset $D_a$ for every $a \in A$;
		\item define the question-answering process: the function $p$ to compute the pertinence of an individual detail $d$ to an archetypal question $q_a$.
	\end{itemize}
	Interestingly, the set of aspects $A$ is task-dependent and must be defined for each explanandum (e.g., manually listing all aspects or automatically extracting the list of aspects from a textual description of the explanation with a tokenizer). Instead, the set of archetypes $Q$, the pertinence function $p$, and the mechanism for extracting $D$ and $D_a$ from $\Phi$ \textit{may always be the same} for all explananda.
	Specifically, the set $A$ of \textit{explanandum aspects} is a collection of (lemmatized) words, and it can be different from the set $I$ of aspects explained by $\Phi$. 
	What is of utmost importance for a $\Phi$ to be a good \textit{explanandum support material} is that $A \subseteq I$.
	
	\subsubsection{Details Extraction and Pertinence Estimation} \label{sec:proposed_solution:practice:details_extraction}
	
	Definition \ref{def:DoX} requires a mechanism to identify the set $D$ of details contained in the \textit{explanandum support material} $\Phi$, as well as a mechanism to identify the sub-sets $D_a \subseteq D$ for every $a \in A$.
	
	A detail $d$ is a snippet of text with some specific characteristics, also called \textit{information unit}. It is a relatively small sequence of words about one or more aspects (i.e., a sub-set of $I$) that is usually extracted from a more complex information bundle (i.e., a paragraph, a sentence). 
	In other terms, these details should carry enough information to describe different parts of an aspect (possibly connected to many other aspects). So we can use them to answer some (archetypal) questions about an $a \in A$ and to correctly estimate a \textit{level of detail}, as required by Definition \ref{def:DoX}.
	
	Considering the characteristics of $D$ and $I$ mentioned above, their most natural representation is a (knowledge) graph.
	A graph is a set of nodes (i.e., $I$) connected by a set of edges (i.e., $D$).
	Therefore, we believe that the simplest way to identify the set of details $D$ may be to extract a graph of \textit{information units} from $\Phi$ on which efficient question-answering could be performed.
	
	The task of answering questions using an extensive collection of documents about diverse topics or from different domains is called open-domain question-answering \cite{chen2020open,DBLP:journals/access/HuangXHWQFZPW20}.
	There are at least three main software architectures for open-domain question-answering: the retriever-reader architecture, the retriever-generator, and the generator-only architecture.
	The first two architectures combine information retrieval techniques and neural reading comprehension or text generation models. In particular, the latter does not involve classical information retrieval, thus being completely end-to-end.
	A famous example of generator-only architecture could be OpenAI's ChatGPT, an adaptive and intelligent dialogue system.
	This type of question-answering algorithm usually relies on huge (i.e., with hundreds of billions of parameters) deep neural networks trained in an unsupervised manner to memorize facts and in a supervised manner to answer questions in a meaningful and coherent way.
	Even though generator-only architectures are capable of impressive results, they tend to write plausible-sounding but incorrect or nonsensical answers.
	One of the reasons for this problem is that this type of architecture is fully end-to-end and needs to perform fact-checking.
	
	
	On the other hand, the retriever-generator and retriever-reader architectures circumvent the latter problem by relying on a system capable of retrieving plausible answers from a knowledge base (or graph) of verified contents.
	The retriever-reader and retriever-generator models usually have an asymptotic time complexity that grows linearly with the number of answers considered for retrieval. That number does not necessarily has to be equal to the number of all the retrievable texts.
	In other words, the time complexity of the answer retrieval system can be intelligently controlled by making it fit the memory and time constraints of a personal computer, e.g., by filtering out all texts unrelated to a question.
	In particular, the retriever-generator rewrites and reprocesses the retrieved information, while the retriever-reader limits itself to extracting it (as it is) and reclassifying it properly.
	While the retriever-generator may still slightly suffer from the problem of generating hallucinated answers, the retriever-reader does not, at the cost of producing less cohesive answers.
	
	For example, a retriever-reader like the one used in \cite{sovrano2021philosophy,sovrano2020legal} for (archetypal) question-answering could be suitable for DoXpy, allowing the identification of meaningful \textit{information units} and also suggesting a mechanism for estimating \textit{pertinence} by extracting from $\Phi$ a graph of $D$ and $I$ designed for answer retrieval.
	Indeed, the aforementioned answer retrieval algorithm consists of a pipeline of AI tools for the extraction from $\Phi$ of a graph of $D$ and $I$ specifically designed to measure the pertinence of $D$ to a set of (archetypal) questions $Q$ on $A$.
	
	More specifically, this graph is extracted by detecting, with a dependency parser, all the clauses within the \textit{explanandum support material} that stand as an edge of the graph.
	In practice, these clauses are represented as special triplets of subjects, templates, and objects called template-triplets.
	Specifically, the templates are composed of the ordered sequence of tokens connecting a subject and an object.
	On the other hand, the subject and the object are represented in these templates by the placeholders \textit{\quotes{\{subj\}}} and \textit{\quotes{\{obj\}}}.
	An example of template-triple is:
	\begin{itemize}
		\item Subject: \textit{\quotes{angina pectoris}}
		\item Template: \textit{\quotes{In particular, \{subj\} happens when some part of your heart does not get enough \{obj\}.}}
		\item Object: \textit{\quotes{oxygen}}
	\end{itemize}
	
	
	Hence, the resulting template-triplets are a sort of function where the predicate is the body, and the object and subject are the parameters. 
	Obtaining a natural language representation (i.e., a detail $d \in D$) of these template-triplets is straightforward by design by replacing the instances of the parameters in the body. 
	This natural representation is then used as a possible answer for retrieval by measuring the (cosine) similarity (or pertinence $p$) between its embedding (obtained through deep language models such as \cite{guo2020multireqa,karpukhin2020dense}) and the embedding of a question $q$.
	
	Notably, as \textit{information units}, \citeauthor{sovrano2021philosophy} \cite{sovrano2021philosophy,sovrano2020legal} use grammatical clauses (meaningful decompositions of grammatical dependency trees) to ensure that the units represent the smallest granularity of information.
	
	As a consequence, using this type of \textit{information units} for \ac{DoX} guarantees:
	\begin{itemize}
		\item a disentanglement of complex information bundles into the most simple units, to correctly estimate the \textit{level of detail} covered by the information pieces, as per Definition \ref{def:DoX};
		\item a better identification of duplicated units scattered throughout the information pieces, to avoid an over-estimation of the \textit{level of detail}.
	\end{itemize}
	All these properties satisfy the requirements that a good detail $d \in D$ should possess for generating a \ac{DoX} score.
	This motivates our decision to use the answer retrieval algorithm of \cite{sovrano2021philosophy,sovrano2020legal} as the main component of the DoXpy pipeline.
	
	\begin{figure}[!htb]
		\centering
		\includegraphics[width=1.\columnwidth]{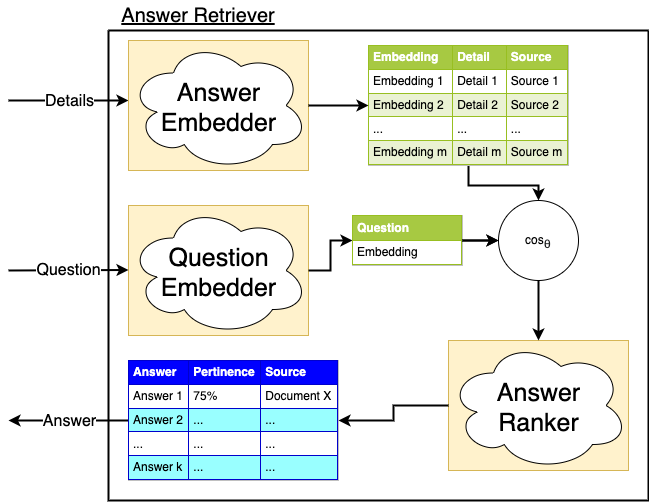}
		\caption{
			\textbf{Flow Diagram of the Answer Retriever used in the pipeline of DoXpy.}
			This figure summarises the answer retrieval algorithm used by DoXpy. 
			A question $q_a$ is given as input to the retriever together with a set of details $D_a$. 
			Then, the details in $D_a$ and $q_a$ are embedded, and their cosine similarity ($\theta$) is used to rank details according to their pertinence to the question.
		}
		\label{fig:yai4hu_answer_retrieval_pipeline}
	\end{figure}
	
	As shown in Figure \ref{fig:yai4hu_answer_retrieval_pipeline}, the retriever-reader used by DoXpy relies on mechanisms for embedding questions and answers in dense numerical representations so that the cosine similarity between the embedding of a question and an answer is a measure of the latter's relevance to the former.
	
	In particular, these embeddings can be obtained (for example) through deep neural networks (i.e., the pertinence function $p$) specialized in answer retrieval and pre-trained on ordinary English to associate similar vectorial representations to a question and its correct answers.
	Examples of these pre-trained deep language models are discussed in the following subsection.
	
	More specifically, let $a$ be the explanandum aspect of a question $q_a$, and $m=<s,t,o>$ be a template-triplet, and $d=t(s,o)$ be the natural language representation of $m$ also called \textit{information unit}, and $z$ the context (i.e., a paragraph, a sentence) from which $m$ was extracted.
	DoXpy performs answer retrieval by retrieving the set $D_a$ of all the template-triplets about $a$ and selecting amongst the natural language representations $d$ of the retrieved template-triplets that are likely to be an answer to $q_a$.
	The probability that $d$ pertinently answers $q_a$ can be estimated as the similarity between the embedding of $d + z$ (i.e., $d$ concatenated with $z$) and the embedding of $q_a$.
	So that if $d + z$ is similar enough to $q_a$, then $z$ is said to be an answer to $q_a$ for \textit{information unit} $d$.
	Therefore, in practice, the algorithm can retrieve an unbounded number of details (i.e., answers).
	
	In particular, a detail is said to be redundant (i.e., duplicated) whenever it contains information that answers an archetypal question $q_a \in Q$ in a manner too similar to that of other (more pertinent) details.
	For example, the detail \quotes{\textit{P is the probability of having a heart disease}} is different but similar to \quotes{\textit{the score P is the probability of having a disease}}.
	However, the former detail is more precise (it speaks of \textit{heart disease} instead of generic diseases) and relevant than the latter in answering the archetypal question \quotes{\textit{What is probability P?}}.
	Therefore, the second detail must be discarded as redundant to prevent DoXpy from considering two details that express the same information differently.
	To do this, DoXpy uses the same deep neural networks used for retrieval to compute the similarity between two answers, discarding those with the lowest relevance scores that share a similarity greater than a threshold $r$.
	
	\begin{figure*}[!tb]
		\centering
		\includegraphics[width=.9\linewidth]{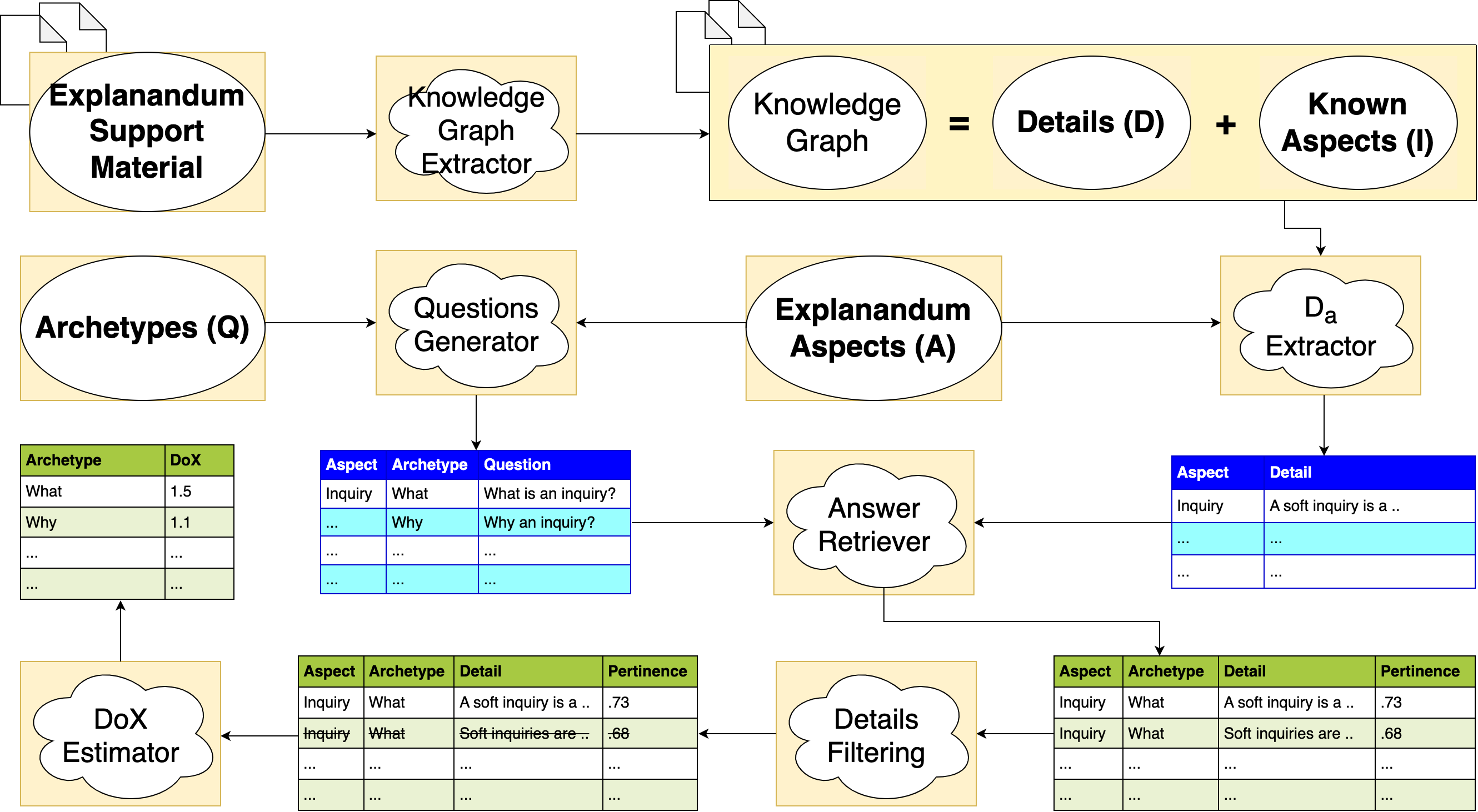}
		\caption{
			\textbf{DoXpy Pipeline.}
			The pipeline starts with extracting a graph from the \textit{explanandum support material} $\Phi$ that is then converted into a set of details $D$. 
			The set of details is then used in combination with the explanandum $A$ and the set of archetypes $Q$ to compute the \ac{DoX}. 
			To do this, we use some deep language models for answer retrieval.
		}
		\label{fig:dox_pipeline}
	\end{figure*}
	
	Consequently, as shown in Figure \ref{fig:dox_pipeline}, the pipeline of DoXpy consists of the following four steps.
	First, a knowledge graph is extracted from the explanandum support material $\Phi$ using the algorithm described in \cite{sovrano2020legal}, thus defining the set of details $D$ and the set of known aspects $I$.
	Secondly, a given set of explanandum aspects $A$ and archetypes $Q$ is used to generate a set of questions $q_a$ for each $a \in A$ and $q \in Q$ and to identify all $D_a \subseteq D$.
	Third, the answer retriever of \cite{sovrano2021philosophy,sovrano2020legal} is used to associate a pertinence score with each $d \in D_a$ for each $q_a$ and (importantly) to identify and filter out duplicate answers.
	Fourth, the formulas in Section \ref{sec:proposed_solution:theory} are used to aggregate the relevance scores and estimate the (average) \ac{DoX} without considering duplicate details.
	
	\subsubsection{Pertinence Functions and Thresholds} \label{sec:proposed_solution:practice:pertinence_functions}
	
	According to Definition \ref{def:DoX}, we need to define a pertinence function $p$ and pick a threshold $t$ to compute the \ac{DoX}.
	As previously discussed, we will use as pertinence function $p$ a deep neural network for answer retrieval.
	The point is that many different deep neural networks exist for this task, i.e., \cite{guo2020multireqa,sovrano2020legal,karpukhin2020dense}, and each one of them has different characteristics producing different pertinence scores.
	So, which model is the right one for computing the \ac{DoX}? Can we use any model?
	
	To answer these questions, we decided to study the behavior of more than one deep language model as pertinence function $p$.
	Assuming that these models get good results on state-of-the-art benchmarks for \textit{pertinence estimation}, we believe that the results of the computation of \ac{DoX} should be consistent across them.
	Hence the models we considered are:
	\begin{itemize}
		\item MiniLM: published by \cite{karpukhin2020dense,reimers2019sentence} and trained on Natural Questions \cite{kwiatkowski2019natural}, TriviaQA \cite{joshi2017triviaqa}, WebQuestions \cite{berant2013semantic}, and CuratedTREC \cite{baudivs2015modeling}.
		\item Multilingual Universal Sentence Encoder: published by \cite{yang2019multilingual} and trained on the Stanford Natural Language Inference corpus \cite{bowman2015large}.
	\end{itemize}
	We experimentally found on the two systems presented in Section \ref{sec:experiments:xai_based_systems} that for both the aforementioned language models, a good pertinence threshold can be $t=0.15$.
	
	\subsubsection{Archetypal Questions} \label{sec:proposed_solution:practice:archetype_selection}
	
	According to Definition \ref{def:DoX}, we need to define a set of archetypal questions $Q$ to compute the explanatory illocution of a snippet of text correctly.
	According to Definition \ref{def:archetypal_question}, an archetypal question is a generic question characterized by one or more interrogative formulas.
	Casting the semantic annotations of individual propositions as narrating an archetypal question-answer pair recently gained increasing attention in computational linguistics \cite{he2015question,fitzgerald2018large,michael2017crowdsourcing,pyatkin2020qadiscourse}, especially in \textit{discourse theory} and the \textit{theory of sentential meaning representations}.
	
	
	On the one hand, \textit{discourse theory} is a branch of linguistics that studies how coherence and cohesion make up a text to form a discourse. 
	So that discourse is said to be coherent if all of its pieces belong together, while it is said to be cohesive if its elements have some common thread. 
	In recent years, many different discourse models have been spelled out, each with different pros and cons.
	Amongst them, we cite the model of the Penn Discourse TreeBank (PDTB for short) \cite{miltsakaki2004penn,prasad2008penn,webber2019penn} because it is considered one of the most generic models of discourse.
	In fact, with little or no change in the model, PDTB appears to be usable for representing discourses of natural languages belonging to different families \cite{zufferey2017annotating}, e.g., Chinese, Arabic, Hindi.
	
	The central assumption behind PDTB is that \quotes{the meaning and coherence of a discourse result partly from how its constituents relate to each other}.
	Specifically, these relations between constituents, called discourse relations, are defined as semantic relations between abstract objects, called elementary discourse units, connected by implicit or explicit connectives, e.g., \quotes{but}, \quotes{then}, \quotes{for example}, \quotes{although}.
	In PDTB, elementary discourse units are spans of text denoting a single event serving as a complete and distinct unit of information that the surrounding discourse may connect to \cite{stede2011discourse}.
	What is of interest to us is that according to Pyatkin et al. \cite{pyatkin2020qadiscourse}, all discourse connectives can be represented as questions:
	\texttt{\begin{inparadesc}
			\item in what manner, 
			\item what is the reason, 
			\item what is the result, 
			\item after what,
			\item what is an example, 
			\item while what, 
			\item in what case,
			\item since when, 
			\item what is contrasted with, 
			\item before what, 
			\item despite what, 
			\item what is an alternative,
			\item unless what,
			\item instead of what, 
			\item what is similar, 
			\item except when, 
			\item until when.
	\end{inparadesc}}
	
	
	On the other hand, the \textit{theories of sentential meaning representation} are grammatical theories that study the relationships between predicates and arguments in a sentence.
	In particular, predicate-argument relationships support answering basic questions such as \textit{who did what to whom}, and they can be captured with models to separate a sentence's meaning from its syntactic representation.
	Amongst these models, we mention the theory of abstract meaning representations \cite{DBLP:conf/acllaw/BanarescuBCGGHK13,DBLP:conf/acl/LangkildeK98}, which can be used to represent whole sentences as (directed and acyclic) graphs of predicates and arguments that can be exploited for tasks such as machine translation (e.g., the conversion of sentences into symbolic knowledge representations, for example, a piece of software written in Prolog that can be used for inference by an automated reasoner), natural language generation and understanding.
	In particular, according to Michael et al., \cite{michael2017crowdsourcing}, all the abstract meaning representations can be encoded as pairs of questions and answers involving the following \textit{archetypes}:
	\texttt{\begin{inparadesc}
			\item what, 
			\item who, 
			\item how, 
			\item where, 
			\item when, 
			\item which, 
			\item whose, 
			\item why.
	\end{inparadesc}}
	
	Interestingly, it is possible to identify a hierarchy or taxonomy of these archetypes, ordered by their intrinsic level of generality or specificity.
	For example, the simplest interrogative formulas, such as those used by the theory of abstract meaning representations, can be seen as the most generic archetypes since they consist of only one interrogative particle: \texttt{what}, \texttt{why}, \texttt{when}, \texttt{who}, etc.
	We will refer to these archetypes as the \textit{primary} ones.
	On the other hand, the archetypes used by the PDTB model (e.g., \texttt{what is the reason}, \texttt{what is the result}) are more complex and specific, building over the primary archetypes.
	For this reason, we will refer to them as \textit{secondary} archetypes.
	
	Even though many more archetypes could be devised (e.g., \texttt{where to} or \texttt{who by}), we believe that the list of questions we provided earlier is already rich enough to be generally representative of any other question, whereas more specific questions can always be framed by using the interrogative particles we considered (e.g., \texttt{why}, \texttt{what}).
	\textit{Primary archetypes} can be used to represent any fact and abstract meaning \cite{bos2016expressive}. In contrast, the \textit{secondary archetypes} can cover all the discourse relations between them (at least according to the PDTB theory).
	
	
	\section{Evaluation of DoXpy: Experiments and Results} \label{sec:experiments}
	
	In Section \ref{sec:proposed_solution:theory} we argued that the degree of explainability of any collection of text (e.g., the output of an XAI) could be measured in terms of \ac{DoX} on a set of chosen \textit{explanandum aspects}.
	In order to verify this assertion and Hypothesis \ref{hyp:main}, we have to show that there is a strong correlation between our \ac{DoX} and the perceived amount of \textit{explainability}.
	To this end, we devised two experiments using some systems making use of XAI (also called XAI-based systems).
	In particular, with the first experiment, we measure explainability \textit{directly}, while with the second, we perform \textit{indirect} measurements obtained through user studies with human subjects.
	
	Measuring explainability \textit{directly} is not possible without a metric like the one we propose (\ac{DoX}), except for a few naive cases.
	One of these cases is undoubtedly when a simple XAI-based system is considered. 
	In fact, in a standard XAI-based system, the amount of \textit{explainability} is (by design) clearly and explicitly dependent on the output of the underlying \ac{XAI}, for the black-box not being explainable by nature. 
	Thus, by masking the output of the \ac{XAI}, the overall system can be forced to be not explainable enough.
	This characteristic can be used to partially verify Hypothesis \ref{hyp:main}, but not in a generic way because this type of verification is based on a comparison with a total lack of explainability and not with different degrees of it.
	
	This is why we decided to measure explainability also \textit{indirectly} with a second experiment, to understand whether \ac{DoX} correlates with the expected effects of explainability on human subjects.
	In other terms, we have to compare \ac{DoX} to existing metrics for explainability based on Cognitive Science (e.g., usability, effectiveness) as shown in Table \ref{tab:literature_comparison}.
	
	If Hypothesis \ref{hyp:main} is correct, the lower the \ac{DoX} score, the fewer explanations can be extracted, and the less effective (as per ISO 9241-210) an explainee is likely to be in achieving explanatory goals that are not covered by the explanations.
	More specifically, \textit{effectiveness} here is defined as \quotes{accuracy and completeness with which users achieve specified goals}. In our case, it is measured through multiple-choice domain-specific quizzes.
	
	We expect an increment in \ac{DoX} always corresponds to an increment in effectiveness, at least on those tasks covered by the information provided by the increment of \ac{DoX}.
	To show this, we borrowed the results of two independent user studies \cite{sovrano2021philosophy,sovrano2022explanatory}, observing how \ac{DoX} correlates with the effectiveness scores measured by these studies. 
	
	\subsection{XAI Systems Considered for the Experiments} \label{sec:experiments:xai_based_systems}
	
	The two systems making use of XAI that we considered as case studies are:
	\begin{itemize}
		\item a heart disease predictor based on XGBoost \cite{chen2016xgboost} and TreeSHAP \cite{lundberg2020local}, concerning healthcare;
		\item a credit approval system based on a simple Artificial Neural Network and CEM \cite{dhurandhar2018explanations}, concerning finance.
	\end{itemize}
	Both these systems are an example of normal \textit{XAI-based explainer}, a one-size-fits-all explanatory mechanism providing the bare output of the \ac{XAI} as a fixed explanation for all users, together with the output of the wrapped AI, a few extra details to ensure the readability of the results, and a minimum of context.
	
	\subsubsection{Finance: the Credit Approval System}
	
	The credit approval system is the same also used by \cite{sovrano2021philosophy,sovrano2022generating}, and it has been designed by IBM to showcase AIX360\footnote{\url{https://aix360.mybluemix.net/explanation_cust}}.
	In particular, this credit approval system uses an artificial neural network to predict a customer's credit risk (and thus decide whether to approve a loan) together with an XAI (called CEM \cite{dhurandhar2018explanations}) to provide post-hoc (static) explanations of the neural network's predictions.
	These explanations aim at helping customers understand whether they have been treated fairly, providing insights into ways to improve their qualifications so as the likelihood of future acceptance can be increased.
	
	A typical use case of this system is the following.
	A customer (e.g., John) applies for a loan from the bank.
	The bank collects sufficient information about the customer. It transmits it to the artificial neural network, which uses it to work out how likely the customer is to repay the loan.
	If the customer's credit risk is low, the loan application is approved, but if the credit risk is too high, the system uses the CEM to explain why.
	
	The artificial neural network behind this credit approval system is trained on the \quotes{FICO HELOC} dataset\footnote{\url{https://fico.force.com/FICOCommunity/s/explainable-machine-learning-challenge?tabset-3158a=a4c37}}, containing anonymized information about loan applications made by real homeowners, to answer the following question: \quotes{What is the decision on the loan request of applicant X?}.
	
	Given the specific characteristics of this credit approval system, we assume that its users' main goal is to understand the causes behind a loan rejection and what to do to get a loan accepted. 
	This is why CEM is deployed to answer the following questions: 
	\begin{itemize}
		\item What are the factors to consider to change the result of the application of applicant X?
		\item How should factor F be modified to change the result of the application of applicant X?
		\item What is the relative importance of factor F in changing the result of the application of applicant X?
	\end{itemize}
	Nonetheless, many other relevant questions might be answered before the user is satisfied and reaches his/her objective. 
	These questions may be: \quotes{How to perform those minimal actions?}, \quotes{Why are these actions so important?}, etc.
	
	Indeed, interpreting the internal parameters and complex calculations of an AI model such as this credit approval system is complicated.
	For example, a layperson trying to obtain a loan might undoubtedly be interested to know that her/his application was rejected (by the AI) mainly due to a high number of credit inquiries on his/her accounts (as CEM can tell). However, this information alone might not be sufficient to achieve her/his goals. 
	These objectives may be beyond the reach of the AI, such as understanding how to effectively reduce the number of inquiries to obtain the loan, what type of credit inquiries may affect his status and the difference between a hard and a soft inquiry.
	
\begin{figure*}[!tb]
		\centering
		\includegraphics[width=.85\linewidth]{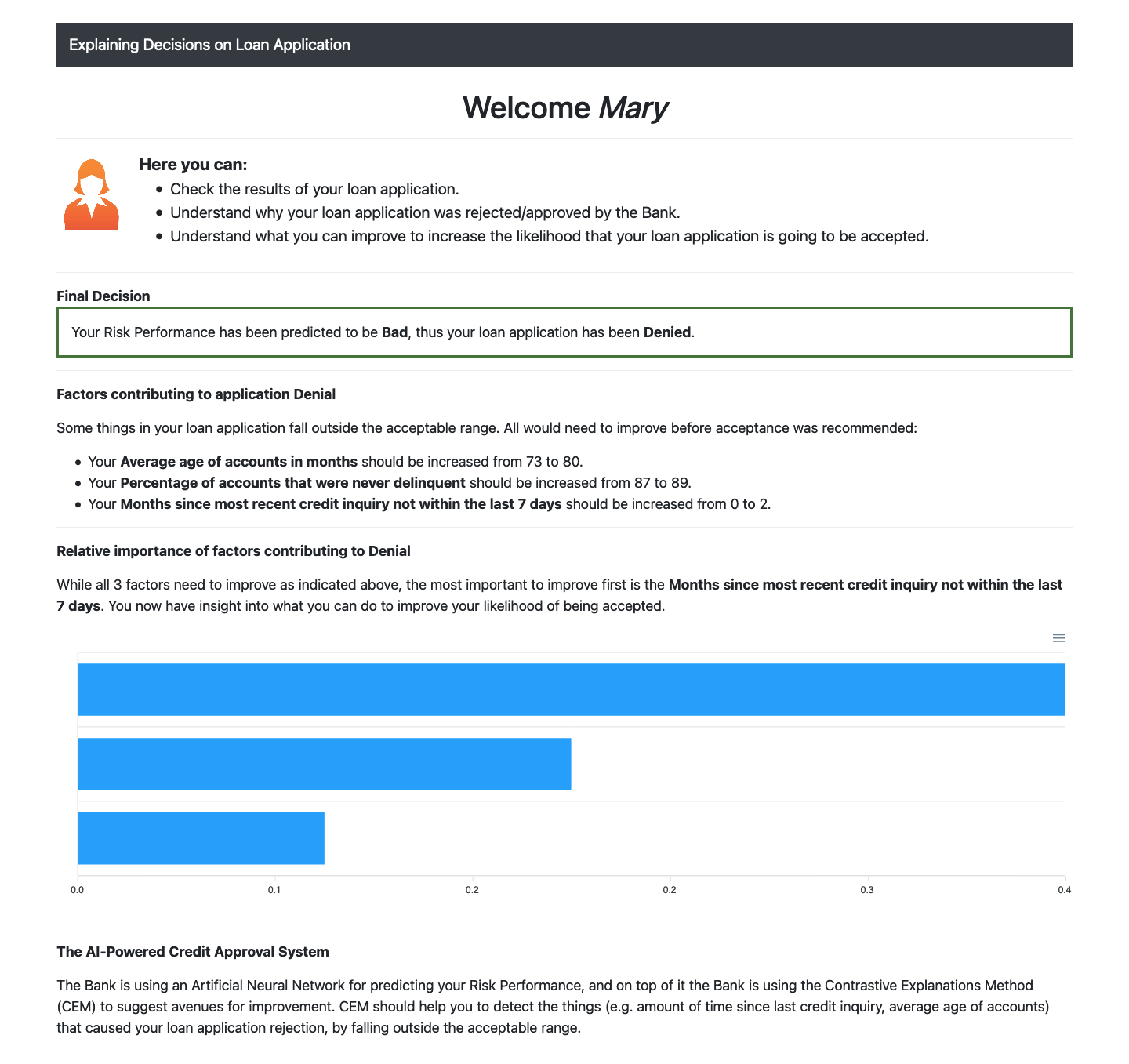}
		\caption{\textbf{Screenshot of the credit approval system.}}
		\label{fig:nxe_ca}
	\end{figure*}

	To summarise, the output of the credit approval system is composed by:
	\begin{itemize}
		\item Context: a titled heading section kindly introducing the user to the system.
		\item AI Output: the decision taken by the artificial neural network for the loan application (i.e., \quotes{denied} or \quotes{accepted}).
		\item \ac{XAI} Output: a section showing the output of the CEM. This output consists of a minimally ordered list of factors deemed to be the most important to change for the outcome of the artificial neural network to be different.
	\end{itemize}
	A screenshot of a web application implementing this credit approval system is shown in Figure \ref{fig:nxe_ca}.
	
	\subsubsection{Healthcare: the Heart Disease Predictor}
	
	Similarly to the credit approval system, the heart disease predictor comes from \cite{sovrano2022generating}.
	In particular, the explanandum of the heart disease predictor is about health, and the system is used by a first-level responder of a help desk for heart disease prevention. 
	More specifically, a first-level responder is responsible for handling the requests for assistance of a patient, forwarding them to the correct physician in the eventuality of a reasonable risk of heart disease.
	
	First-level responders get basic questions from callers, they are not doctors, but they have to decide on the fly whether the caller should speak to a real doctor. 
	So, they quickly use the heart disease predictor to determine what to answer the callers and the subsequent actions to suggest. 
	In other words, this system is used directly by the responder and indirectly by the caller through the responder. 
	These two types of users have different but overlapping goals and objectives. 
	It is reasonable to assume that the responders' goal is to answer the questions of a caller in the most efficient and effective way.
	
	The considered heart disease predictor uses an AI called XGBoost \cite{chen2016xgboost} to predict the likelihood of a patient having a heart disease given its demographics (gender and age), health (diastolic blood pressure, maximum heart rate, serum cholesterol, presence of chest-pain, etc.) and the electrocardiographic (ECG) results. 
	This likelihood is classified into three different risk areas: low (probability $p$ of heart disease below $0.25$), medium ($0.25 < p < 0.75$), or high.
	Therefore, XGBoost is used to answer the following questions: 
	\begin{itemize}
		\item How likely is it that patient X has heart disease?
		\item What is the risk of heart disease for patient X?
		\item What is the recommended action for patient X to treat or prevent heart disease?
	\end{itemize}
	In particular, the dataset used to train XGBoost is the \quotes{UCI Heart Disease Data} \cite{detrano1989international,alizadehsani2019database}.
	
	On top of XGBoost, the heart disease predictor uses TreeSHAP \cite{lundberg2020local}, a famous \ac{XAI} algorithm specialized in tree ensemble models (e.g., XGBoost) for post-hoc explanations.
	In particular, TreeSHAP is used to understand the contribution of each input feature to the output of XGBoost. 
	Therefore, TreeSHAP is used to answer the following questions: 
	\begin{itemize}
		\item What would happen if patient X had factor Y (e.g., chest pain) equal to A instead of B?
		\item What are the most important factors contributing to the predicted likelihood of heart disease for patient X?
		\item How factor Y contributes to the predicted likelihood of heart disease for patient X?
	\end{itemize}
	However, many other important questions should be answered. These include \quotes{What is the easiest thing the patient could do to change his heart disease risk from medium to low?}, \quotes{How could the patient avoid raising one of the factors, preventing his heart disease risk from raise?}.
	
\begin{figure*}[!tb]
		\centering
		\includegraphics[width=.85\linewidth]{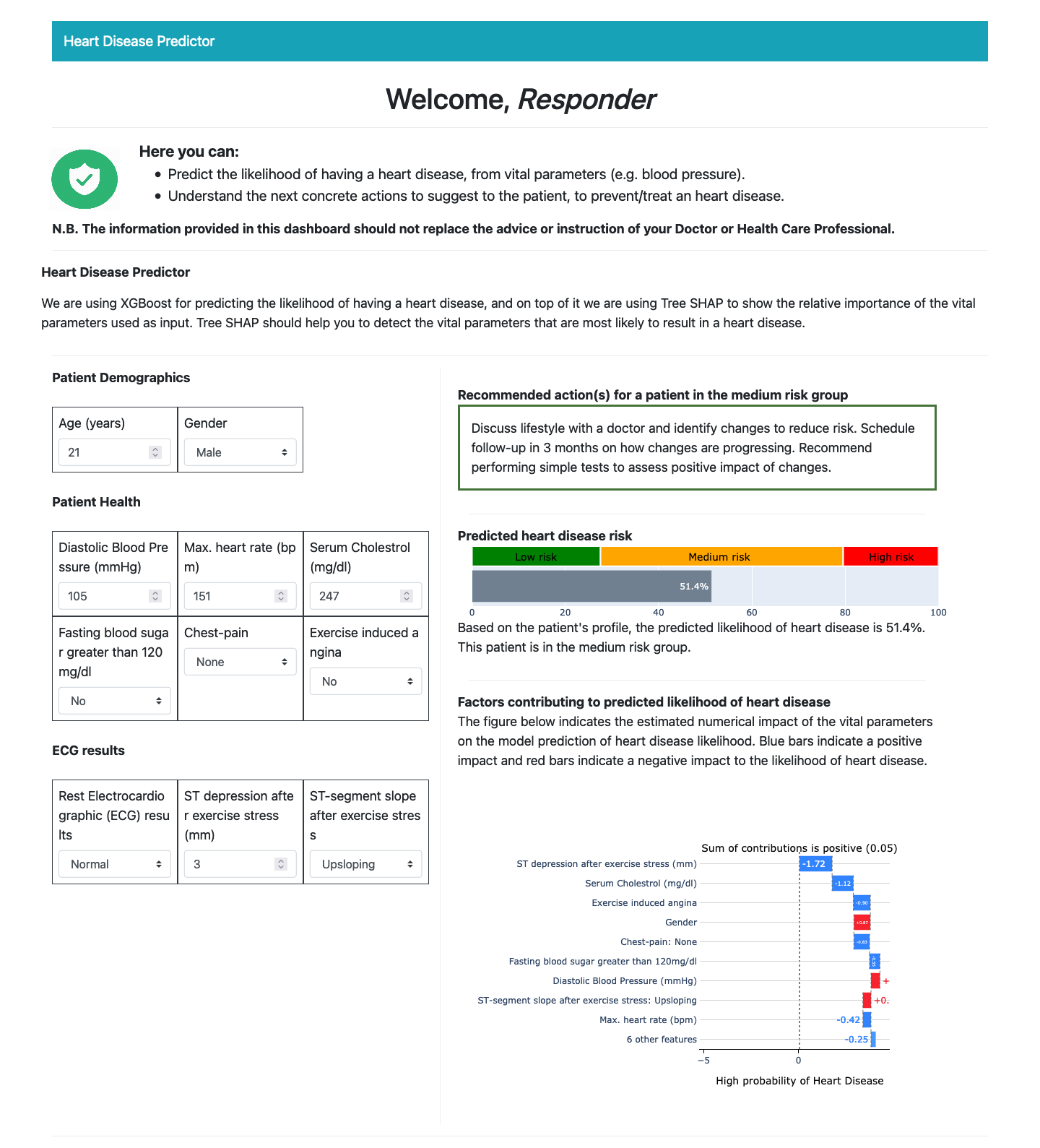}
		\caption{\textbf{Screenshot of the heart disease predictor.}}
		\label{fig:nxe_hd}
	\end{figure*}

	Finally, to summarise, the output of the heart disease predictor is composed by:
	\begin{itemize}
		\item Context: a titled heading section kindly introducing the responder (the user) to the system.
		\item AI Inputs: a panel for entering the patient's biological parameters.
		\item AI Outputs: a section displaying the likelihood of heart disease estimated by XGBoost and a few generic suggestions about the next actions to take.
		\item \ac{XAI} Outputs: a section showing each biological parameter's contribution (positive or negative) to the likelihood of heart disease generated by TreeSHAP.
	\end{itemize}
	A screenshot of a web application implementing this heart disease predictor is presented in Figure \ref{fig:nxe_hd}.
	
	\subsection{1st Experiment: Direct Evaluation on Normal XAI-generated Explanations} \label{sec:experiments:exp1}
	
	In order to verify Hypothesis \ref{hyp:main}, we have to show that there is a strong correlation between \ac{DoX} and the perceived amount of \textit{explainability}.
	To this end, we devised two experiments.
	
	The 1st experiment is meant to shed more light on how a few changes to the explainability of a system affect the estimated \ac{DoX}.
	Specifically, XAI-based systems are considered for this experiment because their amount of \textit{explainability} is, by design, clearly and explicitly dependent on the output of the underlying \ac{XAI}. 
	So, by masking the output of the \ac{XAI}, the overall system can be forced to be less explainable.
	Hence, this characteristic can be exploited to verify the hypothesis in a straightforward but effective way.
	
	In other words, a \ac{XAI}-based system is composed of a black-box AI system wrapped by a \ac{XAI}.
	So, with this experiment, we compare the \ac{DoX} of a normal XAI-based explainer with that of the same system without the \ac{XAI}, also called \textit{normal AI-based explainer}.
	As a result, we expect the (average) \ac{DoX} of the XAI-based explainer to be higher than its wrapped AI.
	
	For this experiment, we used the XAI-based systems defined in Section \ref{sec:experiments:xai_based_systems}.
	Therefore, by simply removing the output of the \ac{XAI} (respectively CEM and TreeSHAP) from these systems, it is possible to obtain the \textit{normal AI-based explainers} we need.
	
	In order to compute the (average) \ac{DoX} of these systems, we take as a set of \textit{explanandum aspects} those targeted by the credit approval system and the heart disease predictor.
	More precisely, the main \textit{explanandum aspects} $A$ targeted by XGBoost \cite{chen2016xgboost} and TreeSHAP \cite{lundberg2020local} in the heart disease predictor are five:
	\begin{itemize}
		\item the recommended action for patient $X$;
		\item the most important factors $Y$ that contribute to predicting the likelihood of heart disease;
		\item the likelihood of heart disease;
		\item the risk $R$ of having a heart disease;
		\item the contribution of $Y$ to predict the likelihood of heart disease for patient $X$.
	\end{itemize}
	While the main \textit{explanandum aspects} $A$ targeted by the Artificial Neural Network and CEM \cite{dhurandhar2018explanations} in the credit approval system are four:
	\begin{itemize}
		\item the factors $F$ to consider for changing the result;
		\item the relative importance of factors $F$ in changing the result;
		\item the risk performance of applicant $X$;
		\item the result of the application of applicant $X$.
	\end{itemize}
	Eventually, after properly converting the images produced by the \textit{XAI-based explainers} to textual explanations, the resulting \textit{explanandum aspects coverage} (i.e., the ratio of ${\lvert A \cap I \lvert}$ to ${\lvert A \lvert}$) of both the heart disease predictor and the credit approval system is 100\%. In contrast, that of their \textit{AI-based explainers} is 48\% and 43\% respectively.
	
	
	\begin{table}[!htb]
		\caption[Results of the 1st Experiment on DoXpy]{
			\textbf{Results of the 1st Experiment on DoXpy\footnotemark.}
			In this table, \ac{DoX} and average (Avg) \ac{DoX} are shown for the credit approval system (CA) and the heart disease predictor (HD). 
			As columns, we have the normal AI-based explainers (AI, for short) and the normal XAI-based explainers (XAI, for short). 
			As rows, we have different explainability estimates using MiniLM (ML) and the Universal Sentence Encoder (TF). 
			For simplicity, for \ac{DoX}, we show only the \textit{primary archetypes}.
		} \label{tab:exp1_dox}
		\centering
		\resizebox{\linewidth}{!}{
			\arrayrulecolor{black}
			\begin{tabular}{|l|l|l|l||l|l|} 
				\hhline{~~--|t|--|}
				\multicolumn{2}{l|}{\multirow{2}{*}{}}                                                                 & \multicolumn{2}{l||}{{\cellcolor[rgb]{0.937,0.937,0.937}}CA}                                                                                                                                                                                                                                                           & \multicolumn{2}{l|}{{\cellcolor[rgb]{0.937,0.937,0.937}}HD}                                                                                                                                                                                                                                                      \\ 
				\hhline{~~--||--|}
				\multicolumn{2}{l|}{}                                                                                  & {\cellcolor[rgb]{0.937,0.937,0.937}}AI                                                                                                                        & {\cellcolor[rgb]{0.937,0.937,0.937}}XAI                                                                                                                & {\cellcolor[rgb]{0.937,0.937,0.937}}AI                                                                                                                 & {\cellcolor[rgb]{0.937,0.937,0.937}}XAI                                                                                                                 \\ 
				\hline
				{\cellcolor[rgb]{0.937,0.937,0.937}}                          & {\cellcolor[rgb]{0.937,0.937,0.937}}ML & 0.46                                                                                                                                                          & 1.52                                                                                                                                                   & 0.53                                                                                                                                                   & 1.49                                                                                                                                                    \\ 
				\hhline{|>{\arrayrulecolor[rgb]{0.937,0.937,0.937}}->{\arrayrulecolor{black}}---||--|}
				\multirow{-2}{*}{{\cellcolor[rgb]{0.937,0.937,0.937}}Avg DoX} & {\cellcolor[rgb]{0.937,0.937,0.937}}TF & 0.23                                                                                                                                                          & 0.86                                                                                                                                                   & 0.27                                                                                                                                                   & 0.84                                                                                                                                                    \\ 
				\hline
				{\cellcolor[rgb]{0.937,0.937,0.937}}                          & {\cellcolor[rgb]{0.937,0.937,0.937}}ML & \begin{tabular}[c]{@{}l@{}}"what": 0.49\\ "how": 0.48\\ "which": 0.47\\ "who": 0.47\\ "why": 0.46\\ "whose": 0.45\\ "when": 0.45\\ "where": 0.44\end{tabular} & \begin{tabular}[c]{@{}l@{}}"which": 1.61\\"how": 1.60\\"what": 1.59\\"why": 1.53\\"when": 1.51\\"where": 1.46\\"who": 1.45\\"whose": 1.41\end{tabular} & \begin{tabular}[c]{@{}l@{}}"why": 0.60\\"which": 0.55\\"what": 0.54\\"how": 0.54\\"whose": 0.49\\"when": 0.47\\"where": 0.47\\"who": 0.46\end{tabular} & \begin{tabular}[c]{@{}l@{}}"why": 1.63\\"which": 1.60\\"what": 1.52\\"how": 1.52\\"whose": 1.51\\"who": 1.40\\"when": 1.38\\"where": 1.38\end{tabular}  \\ 
				\hhline{|>{\arrayrulecolor[rgb]{0.937,0.937,0.937}}->{\arrayrulecolor{black}}---||--|}
				\multirow{-2}{*}{{\cellcolor[rgb]{0.937,0.937,0.937}}DoX}     & {\cellcolor[rgb]{0.937,0.937,0.937}}TF & \begin{tabular}[c]{@{}l@{}}"what": 0.26\\"when": 0.24\\"which": 0.22\\"how": 0.19\\"where": 0.18\\"who": 0.18\\"why": 0.16\\"whose": 0.15\end{tabular}        & \begin{tabular}[c]{@{}l@{}}"what": 0.94\\"when": 0.87\\"which": 0.77\\"where": 0.73\\"how": 0.73\\"who": 0.68\\"why": 0.64\\"whose": 0.55\end{tabular} & \begin{tabular}[c]{@{}l@{}}"what": 0.30\\"when": 0.25\\"which": 0.23\\"who": 0.22\\"how": 0.22\\"where": 0.22\\"why": 0.22\\"whose": 0.18\end{tabular} & \begin{tabular}[c]{@{}l@{}}"what": 0.97\\"when": 0.78\\"how": 0.74\\"which": 0.72\\"who": 0.68\\"where": 0.68\\"why": 0.66\\"whose": 0.56\end{tabular}  \\
				\hline
			\end{tabular}
		}
	\end{table}
	\footnotetext{The numerical values in this table are different from those reported in \cite{sovrano2022how} because we used DoXpy v3.0 instead, which includes several improvements in the information retrieval algorithm that prevent details duplication, as described in Section \ref{sec:proposed_solution:practice}.}
	
	By calculating the \ac{DoX} through DoXpy, we obtained the results shown in Table \ref{tab:exp1_dox}.
	As expected, for both the heart disease predictor and the credit approval system, the experiment results indicate that the (average) \ac{DoX} of all XAI-based explainers is significantly higher than that of AI-based explainers, regardless of the \textit{deep language model} adopted.
	Although, we can see that MiniLM and the Universal Sentence Encoder (the two adopted \textit{language models}) produce comparable but slightly different \ac{DoX} scores, suggesting that the choice of the pertinence function $p$ could sensibly impact the value of \ac{DoX}.
	
	Considering that in this first experiment, we arbitrarily chose a simple set of \textit{explanandum aspects}, what would happen if we considered different and more complex explananda and explanatory contents?
	Furthermore, the result of this experiment is based on comparing the \ac{DoX} of an unexplained system (i.e., the normal \textit{AI-based explainers}) with that of a more explainable system, and this is an exceptional and naive case to consider.
	Therefore, to thoroughly test Hypothesis \ref{hyp:main}, we must understand whether \ac{DoX} behaves as expected even when explainability is present in different and non-zero quantities.
	To this end, explainability can be measured \textit{indirectly} by studying the effectiveness of the resulting explanations on human subjects, as shown in Section \ref{sec:experiments:exp2}.
	
	\subsection{2nd Experiment: A Study of the Effects of Explainability on Human Subjects} \label{sec:experiments:exp2}
	
	The second experiment aims to show whether there is a correlation between \ac{DoX} and the effects of explainability on human subjects.
	We have that a higher explainability implies a greater capacity to explain, hence a greater number of explanations. 
	In short, the lower the \ac{DoX}, the fewer explanations can be produced, and the less effective the explainer is in explanandum-related tasks.
	
	So, if Hypothesis \ref{hyp:main} were true, an increase in the \ac{DoX} of the (explanatory) system would always correspond to a proportional increase of its effectiveness, at least on those tasks covered by the information provided by the increment of \ac{DoX}.
	Therefore, to verify this point, we borrowed two user studies published by \cite{sovrano2021philosophy} and \cite{sovrano2022explanatory}, involving more than 190 human subjects.
	
	Notably, these user studies considered the same explanandum support materials and AI systems used during the first experiment and described throughout Section \ref{sec:experiments:xai_based_systems}, analyzing the effectiveness of explanations given by other explainers when changing the \textit{explanandum support material} and the way explanations are presented to the explainee.
	
	The effectiveness of explanations was measured by giving the explanations to the participants of the user studies and asking them questions to see whether the given explanations helped them to understand the explananda.
	In particular, two domain-specific multiple-choice quizzes (one per explanandum) were used to measure effectiveness, each consisting of questions representing plausible information goals for the system's users. 
	Being impossible and unfeasible to identify all the possible questions a real user would ask to reach its goals, only a few representative questions were considered for the sake of the study.
	It appears from preliminary studies, such as the one by Liao et al. \cite{liao2020questioning}, that users are interested in asking a variety of different questions about an AI-based system, pointing to complex and heterogeneous needs for explainability that go beyond the output of a single \ac{XAI}.
	
	\subsubsection{1st User Study} \label{sec:experiments:exp2:1st_study}
	
	\begin{figure*}[!htb]
		\begin{subfigure}{0.49\linewidth}
			\centering
			\includegraphics[width=1\linewidth]{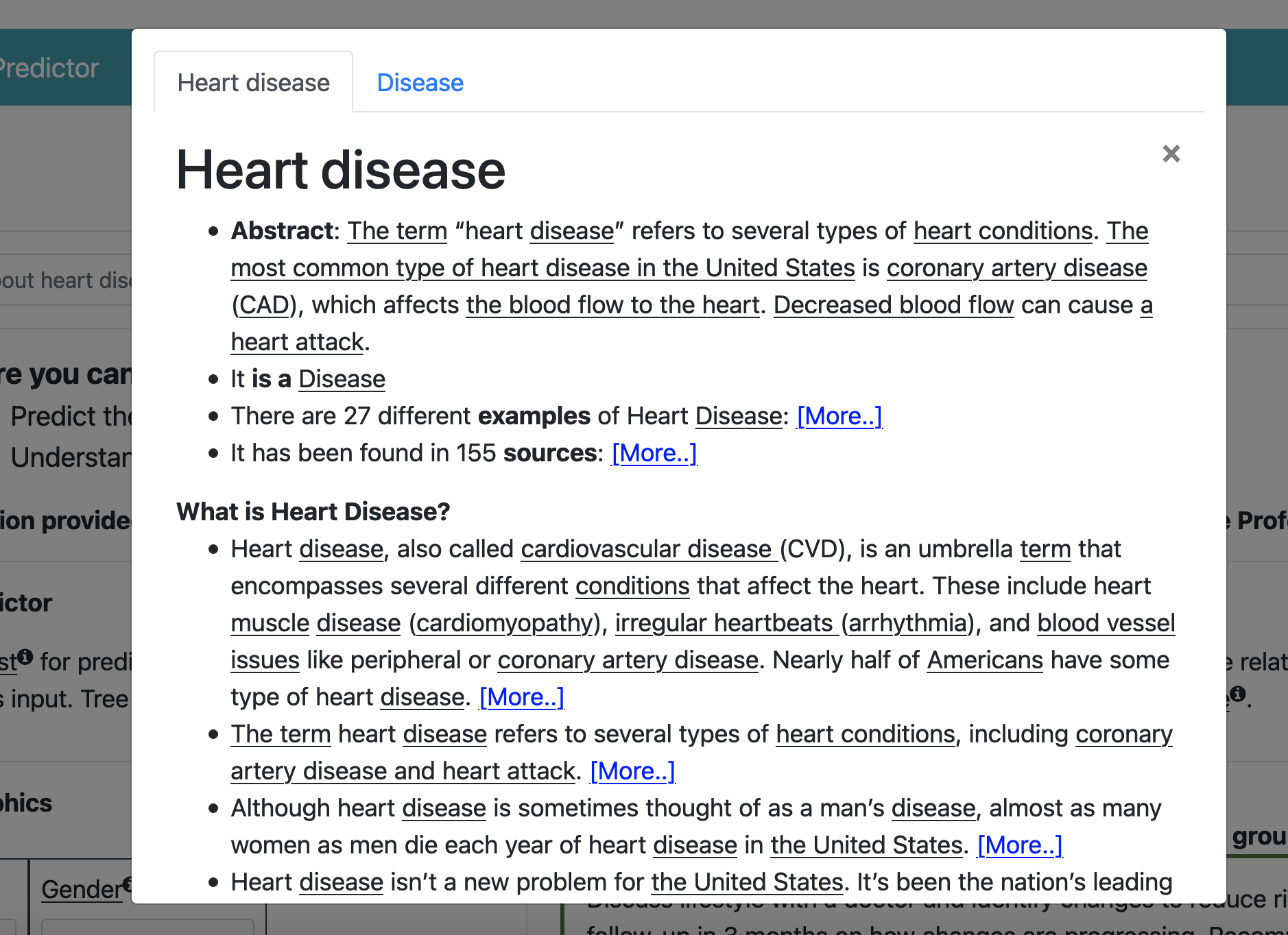}
		\end{subfigure}%
		~ 
		\begin{subfigure}{0.49\linewidth}
			\centering
			\includegraphics[width=1\linewidth]{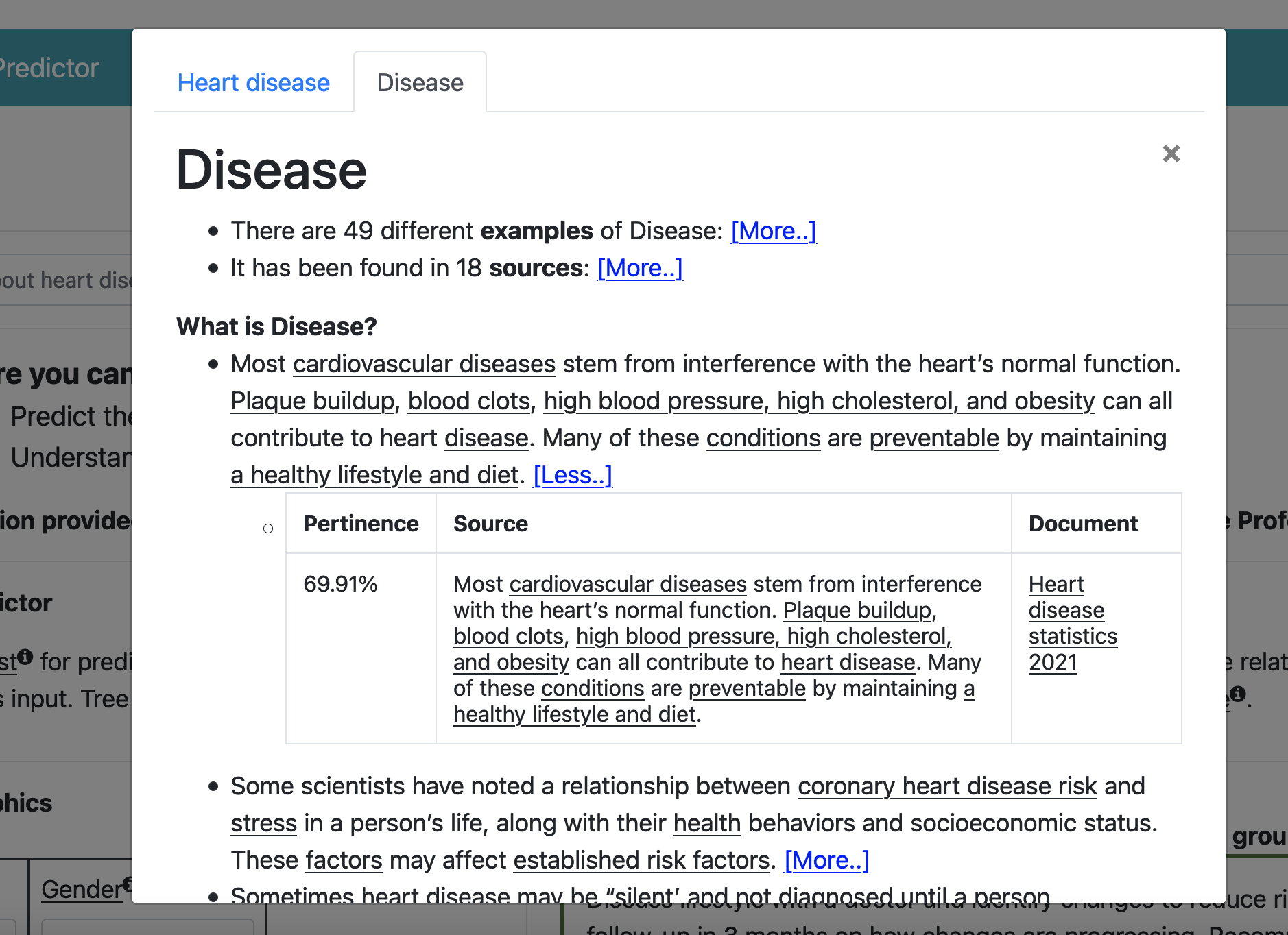}
		\end{subfigure}
		\caption{
			\textbf{Example of Overview}: this figure shows an example of interactive \textit{overview} displaying relevant information about important concepts for the heart disease predictor.
			Clicking on any underlined word would open a new \textit{overview} in a new tab, as shown.
			Furthermore, every given answer is linked to its source document.
		} \label{fig:overview_modal_hd}
	\end{figure*}
	
	On the one hand, the first user study comes from \cite{sovrano2021philosophy}, where a novel mechanism, called \textit{overview-based explainer}, is used to explain extensive collections of heterogeneous documents (i.e., more than 50 web pages) about the credit approval system, in a user-centered and interactive way.
	This is done by organizing knowledge as a graph of explanandum aspects whose related explanations are ordered by relevance and simplicity according to a set of pre-defined archetypal questions (e.g., \texttt{what}, \texttt{how}, \texttt{when}, \texttt{why}).
	In particular, a user can carry out \textit{overviewing} from the initial explanation shown in Figure \ref{fig:nxe_ca} by clicking on the annotated words for which an explanation is needed.
	An example of \textit{overview} is shown in Figure \ref{fig:overview_modal_hd}.
	
	The external resources used by the \textit{overview-based explainer} for the credit approval system consist of 58 webpages, 50 of which come from the website of MyFICO\footnote{\url{https://www.myfico.com}}, while the remaining come from Forbes\footnote{\url{https://www.forbes.com}}, Wikipedia, AIX360\footnote{\url{http://aix360.mybluemix.net}}, and BankRate\footnote{\url{https://www.bankrate.com}}.
	
	\begin{table}[!htb]
		\centering
		\caption{
			\textbf{Quiz of the Credit Approval System.}
			This table contains the quiz used for evaluating the effectiveness of tools explaining the credit approval system.
			In this table, XAI is the normal \textit{XAI-based explainer} (i.e., the webpage shown in Figure \ref{fig:nxe_ca}) and \quotes{OBE} is the overview-based explainer.
			Column \textit{Steps} indicates the minimum number of steps (in terms of links to click, overviews to open, or questions to pose) required by each explanatory tool to provide the correct answer. 
			Negative \textit{steps} means that the correct answer cannot be found. In contrast, 0 \textit{steps} means that the answer is immediately available in the initial explanans (i.e., the content of the webpage shown in Figure \ref{fig:nxe_ca}).
			Column \quotes{Archetype} indicates which interrogative particles represent the question. Many questions are polyvalent in that they can be rewritten using different archetypes.
		}
		\label{tab:ca_quiz}
		\resizebox{\linewidth}{!}{
			\arrayrulecolor{black}
			\begin{tabular}{|p{0.6\linewidth}|l|l|l|l|l|} 
				\hline
				\rowcolor[rgb]{0.937,0.937,0.937} {\cellcolor[rgb]{0.937,0.937,0.937}}                                                                                           & {\cellcolor[rgb]{0.937,0.937,0.937}}                                     & \multicolumn{2}{l|}{\textbf{Steps}}      \\ 
				\hhline{|>{\arrayrulecolor[rgb]{0.937,0.937,0.937}}-->{\arrayrulecolor{black}}--|}
				\rowcolor[rgb]{0.937,0.937,0.937} \multirow{-2}{*}{{\cellcolor[rgb]{0.937,0.937,0.937}}\textbf{Question}}                                                        & \multirow{-2}{*}{{\cellcolor[rgb]{0.937,0.937,0.937}}\textbf{Archetype}} & \textbf{XAI} & \textbf{\textbf{OBE}}  \\ 
				\hline
				What did the credit approval system decide for Mary's application?                                                                                               & \texttt{what}, \texttt{how}                                                                & 0            & 0                         \\ 
				\hline
				What is an inquiry (in this context)?                                                                                                                            & \texttt{what}                                                                     & -1           & 1                         \\ 
				\hline
				What type of inquiries can affect Mary's score, the hard or the soft ones?                                                                                       & \texttt{what}, \texttt{how}                                                                & -1           & 1                         \\ 
				\hline
				What is an example of hard inquiry?                                                                                                                              & \texttt{what}                                                                     & -1           & 1                         \\ 
				\hline
				How can an account become delinquent?                                                                                                                            & \texttt{how}, \texttt{why}                                                                 & -1           & 1                         \\ 
				\hline
				Which specific process was used by the Bank to automatically decide whether to assign the loan?                                                                  & \texttt{what}, \texttt{how}                                                                & 0            & 0                         \\ 
				\hline
				What are the known issues of the specific technology used by the Bank (to automatically predict Mary's risk performance and to suggest avenues for improvement)? & \texttt{what}, \texttt{why}                                                                & -1           & 1                         \\
				\hline
			\end{tabular}
		}
	\end{table}
	
	This first user study compares the effectiveness scores of the credit approval system with and without the possibility for users to perform \textit{overviewing} to demonstrate that the \textit{overview-based explainer} generates more effective explanations than the baseline.
	Specifically, effectiveness scores are generated by users interacting with the system and answering a multiple-choice quiz (shown in Table \ref{tab:ca_quiz}) on the credit approval system.
	In particular, each question of the multiple choice quiz has 4 to 8 plausible answers, of which only one is (the most) correct. 
	At the end of the quiz, answers are automatically scored as correct (score 1) or not (score 0), and the resulting scores are added together to form the effectiveness score.
	For example, for the question \quotes{What did the Credit Approval System decide for Mary's application?}, the correct answer is \quotes{It was rejected}, and the wrong answers are \quotes{Nothing} or \quotes{I do not know}.
	
	For this user study, 103 participants were recruited (57 males, 44 females, and two unknowns, ages 18-55) on the online platform Prolific \cite{palan2018prolific}. 
	All the participants were recruited among those who:
	\begin{inparaenum}
		\item are resident in UK, US, or Ireland;
		\item have a Prolific acceptance rate greater or equal to 75\%\footnote{Mainly because they are unlikely to answer poorly/randomly to questions.}.
	\end{inparaenum}
	Participants were randomly assigned to use the credit approval system with or without \textit{overview-based explainer} in a between-subjects test.
	The credit approval system without \textit{overview-based explainer} is also called normal \textit{XAI-based explainer} because it only explains through the output of an XAI.
	
	In the end, 51 participants evaluated the normal \textit{XAI-based explainer}, and 52 evaluated the \textit{overview-based explainer}.
	For more details about this user study, read \cite{sovrano2021philosophy}.
	
	\subsubsection{2nd User Study} \label{sec:experiments:exp2:2nd_study}
	
	The second user study comes instead from \cite{sovrano2022explanatory} and concerns the heart disease predictor.
	Similar to the first, this study compares the effectiveness of an \textit{overview-based explainer} called YAI4Hu and that of a normal \textit{XAI-based explainer}.
	In addition, this second study also investigates the effectiveness of two other explainers: a \textit{two-level explainer} and a \textit{how-why explainer}.
	
	The \textit{two-level explainer} is static, as the normal \textit{XAI-based explainer}. It is made of the output of the XAI (shown in Figure \ref{fig:nxe_hd}) directly connected to a second (non-expandable) layer of information consisting of an exhaustive and verbose set of autonomous static explanatory resources.
	The \textit{two-level explainer} is organized, therefore, as a very long text document (more than 50 pages per system, when printed), structured in titled Sections and prefixed with a table of content with hypertext links.
	
	On the other hand, the \textit{how-why explainer} is like the \textit{overview-based explainer}, but it uses only the archetypes \texttt{why} and \texttt{how} for generating explanations.
	Furthermore, also YAI4Hu is an extension of the \textit{overview-based explainer} that instead adds a mechanism (called \textit{open-ended questioning}) for users to ask their questions to the system.
	More specifically, \textit{open-ended questioning} can be performed by asking questions in English through a search box that uses the graph-based answer retrieval mechanism described in Section \ref{sec:proposed_solution:practice}.
	Importantly, YAI4Hu, the \textit{how-why explainer} and the \textit{two-level explainer} share the same \textit{explanandum support material}.
	
	Such \textit{explanandum support material} is composed by the contents shown in Figure \ref{fig:nxe_hd} and a set of external resources carefully selected to cover the topics of the heart disease predictor that consists of 103 webpages, 75 of which come from the website of the U.S. Centers for Disease Control and Prevention\footnote{\url{https://www.cdc.gov}}, while the remaining from the American Heart Association\footnote{\url{https://www.heart.org}}, Wikipedia, MedlinePlus\footnote{\url{https://medlineplus.gov}}, MedicalNewsToday\footnote{\url{https://www.medicalnewstoday.com}} and other minor sources.
	
	For this second user study, 89 different participants were recruited amongst the university students of the following courses of study\footnote{All the courses of study were of an Italian university, and only the master's degrees were international, i.e., with English teachings and students from countries other than Italy.}:
	\begin{inparadesc}
		\item bachelor's degree in computer science;
		\item bachelor's degree in management for informatics;
		\item master's degree in digital humanities;
		\item master's degree in artificial intelligence.
	\end{inparadesc}
	The 89 participants were randomly allocated for testing only one of the three types of explainers.
	In other words, similarly to the first user study, also this second study followed a between-subjects design.
	In the end, there were approximately 20 participants per explainer.
	
	\begin{table}[!htb]
		\centering
		\caption{
			\textbf{Quiz of the Heart Disease Predictor.}
			This table contains the quiz used for evaluating the effectiveness of tools explaining the heart disease predictor.
			XAI is the normal \textit{XAI-based explainer} (i.e., the webpage shown in Figure \ref{fig:nxe_hd}), HWN is the \textit{how-why explainer}, and 2EC is the \textit{two-level explainer}.
			Column \textit{Steps} indicates the minimum number of steps (in terms of links to click, overviews to open, or questions to pose) required by each explanatory tool to provide the correct answer. 
			Negative \textit{steps} means that the correct answer cannot be found. In contrast, 0 \textit{steps} means that the answer is immediately available in the initial explanans (i.e., the content of the webpage shown in Figure \ref{fig:nxe_hd}).
			Column \quotes{Archetype} indicates which interrogative particles represent the question.
		}
		\label{tab:hd_quiz}
		\resizebox{\linewidth}{!}{
			\arrayrulecolor{black}
			\begin{tabular}{|p{0.6\linewidth}|l|l|l|l|l|} 
				\hline
				\rowcolor[rgb]{0.937,0.937,0.937} {\cellcolor[rgb]{0.937,0.937,0.937}}                                                     & {\cellcolor[rgb]{0.937,0.937,0.937}}                                     & \multicolumn{4}{l|}{\textbf{Steps}}                                                      \\ 
				\hhline{|>{\arrayrulecolor[rgb]{0.937,0.937,0.937}}-->{\arrayrulecolor{black}}----|}
				\rowcolor[rgb]{0.937,0.937,0.937} \multirow{-2}{*}{{\cellcolor[rgb]{0.937,0.937,0.937}}\textbf{Question}}                  & \multirow{-2}{*}{{\cellcolor[rgb]{0.937,0.937,0.937}}\textbf{Archetype}} & \textbf{XAI} & \textbf{\textbf{2EC}} & \textbf{\textbf{HWN}} & \textbf{\textbf{YAI4Hu}}  \\ 
				\hline
				What are the most important factors leading that patient to medium risk of heart disease?                                & \texttt{what}, \texttt{why}                                                                & 0            & 0                     & 0                     & 0 (no OQ)                \\ 
				\hline
				What is the easiest thing the patient could do to change his heart disease risk from medium to low?          & \texttt{what}, \texttt{how}                                                                & 0            & 0                     & 0                     & 0 (no OQ)                \\ 
				\hline
				According to the predictor, what serum cholesterol level is needed to shift the heart disease risk from medium to high? & \texttt{what}, \texttt{how}                                                                & 0            & 0                     & 0                     & 0 (no OQ)                \\ 
				\hline
				How could the patient avoid raising bad cholesterol, preventing his heart disease risk from shifting from medium to high?       & \texttt{how}                                                                      & -1           & 1                     & 2                     & 2                         \\ 
				\hline
				What tests can be done to measure bad cholesterol levels in the blood?                                             & \texttt{what}, \texttt{how}                                                                & -1           & 1                     & -1                    & 1                         \\ 
				\hline
				What are the risks of high cholesterol?                                                                                    & \texttt{what}, \texttt{why not}                                                            & -1           & 1                     & 2                     & 1                         \\ 
				\hline
				What is LDL?                                                                                                               & \texttt{what}                                                                     & -1           & 1                     & 2                     & 1                         \\ 
				\hline
				What is Serum Cholesterol?                                                                                                  & \texttt{what}                                                                     & -1           & 1                     & 1                     & 1                         \\ 
				\hline
				What types of chest pain are typical of heart disease?                                                                     & \texttt{what}, \texttt{how}                                                                & -1           & 1                     & 1                     & 1                         \\ 
				\hline
				What is the most common type of heart disease in the USA?                                                                  & \texttt{what}                                                                     & -1           & 1                     & 1                     & 1                         \\ 
				\hline
				What are the causes of angina?                                                                                             & \texttt{what}, \texttt{why}                                                                & -1           & 1                     & 2                     & 1                         \\ 
				\hline
				What kind of chest pain do you feel with angina?                                                                           & \texttt{what}, \texttt{how}                                                                & -1           & 1                     & 1                     & 1                         \\ 
				\hline
				What are the effects of high blood pressure?                                                                               & \texttt{what}, \texttt{why not}                                                            & -1           & 1                     & 1                     & 1                         \\ 
				\hline
				What are the symptoms of high blood pressure?                                                                              & \texttt{what}, \texttt{why}, \texttt{how}                                                           & -1           & 1                     & 1                     & 1                         \\ 
				\hline
				What are the effects of smoking on the cardiovascular system?                                                              & \texttt{what}, \texttt{why not}                                                            & -1           & 1                     & 3                     & 1                         \\ 
				\hline
				How can the patient increase his heart rate?                                                                               & \texttt{how}                                                                      & -1           & 1                     & 3                     & 1                         \\ 
				\hline
				How can the patient try to prevent a stroke?                                                                               & \texttt{how}                                                                      & -1           & 1                     & 3                     & 2                         \\ 
				\hline
				What is a Thallium stress test?                                                                                            & \texttt{what}, \texttt{why}                                                                & -1           & 1                     & 3                     & 1                         \\
				\hline
			\end{tabular}
		}
	\end{table}
	
	Each participant evaluated the effectiveness of the four explainers by taking the multiple-choice quiz shown in Table \ref{tab:hd_quiz}.
	At the end of the effectiveness quiz, answers were automatically scored as correct (score 1) or not (score 0), and the resulting scores were added together to form the effectiveness score.
	For further details about this user study, read \cite{sovrano2022generating}.
	
	\subsubsection{Results of 2nd Experiment}
	
	\begin{figure}
		\centering
		\includegraphics[width=.5\columnwidth]{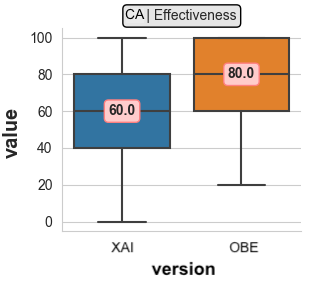}
		\caption{
			\textbf{1st User Study: Effectiveness Scores on Questions that cannot be answered with the information provided by the XAI-based explainer.}
			This figure shows a comparison of the median effectiveness scores obtained on the credit approval system (CA) with the normal \textit{XAI-based explainer} (XAI; the blue one) and YAI4Hu without open-ended question-answering (called \textit{overview-based explainer} or OBE for short; the orange one) on those questions whose answer is not provided by the XAI-based explainer. 
			Results are shown as box plots (25th, 50th, 75th percentile, and whiskers covering all data and outliers). 
			The numerical value of the medians is shown inside pink boxes. 
			Differently from \cite{sovrano2021philosophy}, here effectiveness scores are normalised in $[0,100]$.
		}
		\label{fig:1st_user_study_uncovered}
	\end{figure}
	
	Both the results of the (first) user study involving 89 participants and the (second) user study involving 103 participants indicate that a better (i.e., more explainable) \textit{explanandum support material} implies an explainer capable of producing more effective explanations.
	As also shown in Figure \ref{fig:1st_user_study_uncovered}, according to a one-sided Mann-Whitney U-Test, there is enough statistical evidence to claim that the instance of YAI4Hu considered for the second user study is more effective in credit approval system (U=849.5, \textit{p}=.007) than the \textit{XAI-based explainer} on those questions that cannot be answered by the \ac{XAI} (i.e., questions number 2, 3, 4, 5 and 7 in Table \ref{tab:ca_quiz}).
	
	\begin{figure}
		\centering
		\includegraphics[width=.5\columnwidth]{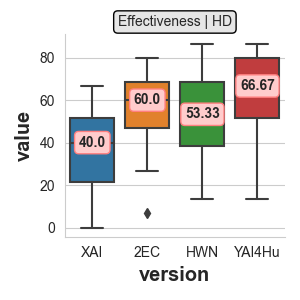}
		\caption{
			\textbf{2nd User Study: Effectiveness Scores on Questions that cannot be answered with the information provided by the XAI-based explainer.}
			Comparison of the results achieved on the heart disease predictor (HD) with the normal \textit{XAI-based explainer} and the other explainers, only on those questions whose aspects are \textit{not covered} by the information presented by the XAI. 
			In particular, the other explainers are the \textit{two-layered explainer} (2EC), the \textit{how-why explainer} (HWN), and YAI4Hu.
			For more details about interpreting this figure, read the caption of Figure \ref{fig:1st_user_study_uncovered}. 
			Effectiveness scores are normalized in $[0,100]$.
		}
		\label{fig:2nd_user_study_uncovered}
	\end{figure}
	
	Moreover, as shown in Figure \ref{fig:2nd_user_study_uncovered}, the same can be said for the heart disease predictor in the first user study.
	As expected, also in this case, we see the median effectiveness score of the normal XAI-based explainer being significantly lower than the other explainers on the questions that the \ac{XAI} cannot answer (i.e., the questions with negative steps in Table \ref{tab:hd_quiz}).
	More precisely, according to some one-sided Mann-Whitney U-tests, there is enough statistical evidence to claim that YAI4Hu is better than the XAI-based explainer (U=40, \textit{p}=.0002) on those questions. The same can be said about the two-level explainer (U=48, \textit{p}=.003) and the how-why explainer (U=65.5, \textit{p}=.02).
	
	Indeed, the difference between a normal XAI-based explainer and the other explainers is twofold. 
	First of all, the explanations produced by YAI4Hu and the how-why explainer are interactive and more user-centered, while those of the normal XAI-based system are not.
	Secondly, the normal XAI-based explainer considers a smaller amount of explainable information.
	YAI4Hu, the how-why explainer, and the two-level explainer produce their explanations using more than 50 extra web pages that the XAI-based explainer does not see.
	This last difference allows us to exploit these user studies to test Hypothesis \ref{hyp:main} further.
	The amount of information the normal XAI-based explainer handles is $\frac{1}{100}$ of all the other explainers.
	
	In order to show that an increment in \ac{DoX} causes a consequent increment in the effectiveness of explanations, we have to compute the \ac{DoX} scores of the normal XAI-based explainer and the \ac{DoX} scores of the other explainers involved in the user study.
	To do so, we identified the set of \textit{explanandum aspects} $A$ from the quizzes used to generate the effectiveness scores (see Table \ref{tab:ca_quiz} and Table \ref{tab:hd_quiz}). 
	These quizzes define what the users should know to be effective, indirectly defining what is essential for the system to explain: the \textit{explanandum aspects}.
	
	Eventually, if Hypothesis \ref{hyp:main} were true, we would expect that the greater \ac{DoX} is, the greater the effectiveness of an explainer.
	Notably, the opposite is not necessarily correct. Two explainers (with different presentation logics; e.g., the \textit{two-layered explainer} and YAI4Hu) might have different effectiveness scores despite having the same \ac{DoX}.
	
	Computing the \ac{DoX} scores for this second experiment, we got the results shown in Table \ref{tab:exp2_dox}.
	Importantly, these results confirmed our expectations for them.
	They indicate that the \textit{two-level explainer}, the \textit{overview-based explainer}, the \textit{how-why explainer}, and YAI4Hu have higher \ac{DoX} scores than the normal \textit{XAI-based explainer} regardless their presentation logic.
	
	\begin{table}[!htb]
		\centering
		\caption{
			\textbf{Results of the 2nd Experiment on DoXpy.}
			The scores in this table are different from those of the first experiment (Table \ref{tab:exp1_dox}) because a different explanandum is considered for the second experiment. 
			In this table, \ac{DoX} and average (Avg) \ac{DoX} are shown for the credit approval system (CA) and the heart disease predictor (HD). 
			As columns, we have the normal XAI-based explainers (XAI, for short) and the other explainers, i.e., YAI4Hu, the two-level explainer, and the how-why narrator. 
			For more details about interpreting this table, read the caption of Table \ref{tab:exp1_dox}.
		}
		\label{tab:exp2_dox}
		\resizebox{\linewidth}{!}{
			\arrayrulecolor{black}
			\begin{tabular}{|l|l|l|l|l|l|} 
				\hhline{~~----|}
				\multicolumn{2}{l|}{\multirow{2}{*}{}}                                                                 & \multicolumn{2}{l|}{{\cellcolor[rgb]{0.937,0.937,0.937}}CA}                                                                                                                                                                                                                                                             & \multicolumn{2}{l|}{{\cellcolor[rgb]{0.937,0.937,0.937}}HD}                                                                                                                                                                                                                                                              \\ 
				\hhline{~~----|}
				\multicolumn{2}{l|}{}                                                                                  & {\cellcolor[rgb]{0.937,0.937,0.937}}XAI                                                                                                                & {\cellcolor[rgb]{0.937,0.937,0.937}}Others                                                                                                                     & {\cellcolor[rgb]{0.937,0.937,0.937}}XAI                                                                                                                & {\cellcolor[rgb]{0.937,0.937,0.937}}Others                                                                                                                      \\ 
				\hline
				{\cellcolor[rgb]{0.937,0.937,0.937}}                          & {\cellcolor[rgb]{0.937,0.937,0.937}}ML & 1.19                                                                                                                                                   & 18.75                                                                                                                                                          & 0.21                                                                                                                                                   & 21.59                                                                                                                                                           \\ 
				\hhline{|>{\arrayrulecolor[rgb]{0.937,0.937,0.937}}->{\arrayrulecolor{black}}-----|}
				\multirow{-2}{*}{{\cellcolor[rgb]{0.937,0.937,0.937}}Avg DoX} & {\cellcolor[rgb]{0.937,0.937,0.937}}TF & 0.72                                                                                                                                                   & 12.86                                                                                                                                                          & 0.16                                                                                                                                                   & 17.55                                                                                                                                                           \\ 
				\hline
				{\cellcolor[rgb]{0.937,0.937,0.937}}                          & {\cellcolor[rgb]{0.937,0.937,0.937}}ML & \begin{tabular}[c]{@{}l@{}}"which": 1.26\\"how": 1.26\\"when": 1.25\\"what": 1.24\\"who": 1.18\\"why": 1.16\\"where": 1.13\\"whose": 1.12\end{tabular} & \begin{tabular}[c]{@{}l@{}}"how": 20.32\\"what": 19.78\\"when": 19.59\\"which": 19.04\\"why": 19.00\\"whose": 17.34\\"where": 17.19\\"who": 17.09\end{tabular} & \begin{tabular}[c]{@{}l@{}}"which": 0.23\\"how": 0.21\\"why": 0.21\\"whose": 0.21\\"what": 0.21\\"when": 0.20\\"where": 0.20\\"who": 0.20\end{tabular} & \begin{tabular}[c]{@{}l@{}}"why": 24.01\\"which": 22.90\\"how": 22.90\\"what": 21.76\\"whose": 21.37\\"when": 21.01\\"where": 19.91\\"who": 19.61\end{tabular}  \\ 
				\hhline{|>{\arrayrulecolor[rgb]{0.937,0.937,0.937}}->{\arrayrulecolor{black}}-----|}
				\multirow{-2}{*}{{\cellcolor[rgb]{0.937,0.937,0.937}}DoX}     & {\cellcolor[rgb]{0.937,0.937,0.937}}TF & \begin{tabular}[c]{@{}l@{}}"what": 0.88\\"when": 0.75\\"how": 0.69\\"which": 0.67\\"who": 0.66\\"where": 0.64\\"why": 0.57\\"whose": 0.55\end{tabular} & \begin{tabular}[c]{@{}l@{}}"what": 15.97\\"when": 13.70\\"how": 12.15\\"who": 11.89\\"where": 11.33\\"which": 11.06\\"why": 9.68\\"whose": 9.32\end{tabular}   & \begin{tabular}[c]{@{}l@{}}"what": 0.19\\"how": 0.16\\"when": 0.15\\"who": 0.15\\"which": 0.14\\"where": 0.14\\"why": 0.14\\"whose": 0.12\end{tabular} & \begin{tabular}[c]{@{}l@{}}"what": 20.67\\"when": 17.42\\"how": 16.45\\"who": 15.95\\"which": 15.90\\"why": 15.30\\"where": 15.25\\"whose": 13.35\end{tabular}  \\
				\hline
			\end{tabular}
		}
	\end{table}
	
	\section{Discussion and Analysis of Empirical Results: How to Use DoX for Assessing Law Compliance} \label{sec:results_discussion_n_limitations}
	
	The results of all experiments and user studies showed that Hypothesis \ref{hyp:main} is valid.
	We see that \ac{DoX} increases whenever a black-box AI is enclosed in a \ac{XAI} and that an increase in \ac{DoX} corresponds to a statistically significant increase in the effectiveness of the explanatory system.
	Therefore, we believe that our technology for estimating the \ac{DoX} might be used for an objective and lawful algorithmic explainability assessment, as soon as what is needed to be explained can be identified under the requirements of the law in the form of a set of precise \textit{explanandum aspects}.
	To guarantee the reproducibility of the experiments, we published the source code of DoXpy\footnote{\url{https://github.com/Francesco-Sovrano/DoXpy}}, as well as the code of the XAI-based systems, the user study questionnaires, and the remaining data mentioned within this paper.
	
	In particular, the results of the first experiment tell us that whenever new information about different aspects to be explained is added to the \textit{explanandum support material}, the \ac{DoX} scores increase, and this is also true when changing the set of \textit{explanandum aspects}, as we did with the second experiment.
	Furthermore, the results of the second experiment tell us that whenever the \ac{DoX} scores increase, the overall effectiveness of the explanations generated from the \textit{explanandum support material} increases as well.
	This is true even for the \textit{two-level explainer}, even though it is not interactive and does not re-organize information to make it simpler and easier to access, dumping on the user dozens of pages of content.
	
	Our user studies involved more than 190 participants and were consistent across two somewhat different and broad user pools, producing statistically significant results (with p-values lower than $0.05$).
	Therefore, considering that \textit{explainability} is fundamentally the \textit{ability to explain}, the two experiments combined tell us that our (average) \ac{DoX} can quantitatively approximate the degree of explainability of information.
	In other words, we conclude from our experiments that \ac{DoX} \textit{can} be used as a proxy for measuring the explainability of an explanatory system, as long as a set of \textit{explanandum aspects} can be defined.
	\ac{DoX} is deterministic and entirely objective, and it could be used as a cheaper alternative to expensive non-deterministic user studies.
	
	We are convinced that \ac{DoX} may have a role in all applications where it is crucial to evaluate explainability objectively.
	Indeed, the main benefit of \ac{DoX} is that it works with any set of \textit{explanandum aspects} $A$. Therefore it can be used to quantify how the explanations given by an AI are aligned with any of the Business-to-Business, and Business-to-Consumer requirements identified by Bibal et al. \cite{bibal2021legal}.
	
	In particular, for each Business-to-Business and Business-to-Consumer requirement we may have the following set of \textit{explanandum aspects} $A$:
	\begin{itemize}
		\item \textit{Providing the main features used in a decision by the AI}: $A$ can be the set of main feature labels used for a decision. This list can be generated with a \ac{XAI} like CEM, TreeSHAP, or others.
		\item \textit{Providing all features processed by the AI}: in this case, $A$ is the set of all the feature labels considered by the AI.
		\item \textit{Providing a comprehensive explanation of a specific decision taken by the AI}: $A$ can be the set of aspects deemed relevant to the decision of the AI, i.e., what is the AI, what are the known issues of the AI, or all the other aspects discussed in \cite{sovrano2020modelling}.
		\item \textit{Providing the underlying logical model followed by the AI}: in this case, $A$ can be the set of all the nouns or noun/verbal phrases used in the textual description of the logical model of the AI.
	\end{itemize}
	Hence, the benefits of using \ac{DoX} over a normal user study are manifold, in fact:
	\begin{itemize}
		\item \ac{DoX} reduces testing costs normally sustained during subject-based evaluations.
		\item \ac{DoX} allows the direct measurement of the degree of explainability of any piece of information for which a meaningful textual representation is written in a natural language (i.e., English).
		\item \ac{DoX} disentangles the evaluation of the \textit{explanandum support material} from that of the explainer (or presentation logic) and the interface.
	\end{itemize}
	In other words, \ac{DoX} is a fully objective metric that could be used to understand whether a piece of information is sufficient to explain something regardless of whether the resulting explanations have been perceived as satisfactory and good by the explainees. 
	We deem this characteristic of \ac{DoX} to be very important: a poor degree of explainability objectively implies poor explanations, no matter how good the adopted explanatory process is (or how it is perceived): \quotes{Users also do not necessarily perform better with systems that they prefer and trust more. To draw correct conclusions from empirical studies, explainable AI researchers should be wary of evaluation pitfalls, such as proxy tasks and subjective measures} \cite{buccinca2020proxy}.
	
	Despite all the good properties supported by both theory and empirical results, we found that \ac{DoX} may have limitations that we plan to address in future works.
	
	First of all, the results of the second experiment show that explanatory systems with the same \ac{DoX} could be usable and effective in different ways.
	Indeed, this points to the fact that \ac{DoX} should not be considered as a total replacement to user studies but rather as a cheaper alternative to consider while developing complex explanatory systems.
	In other words, \ac{DoX} cannot fully replace subjective metrics (i.e., usability) if one wants to evaluate the user-centrality of an explanatory system or interface. 
	On the other hand, \ac{DoX} is probably better than subjective metrics if one wants to objectively evaluate the contents of an explanatory system to understand how many questions can be adequately answered: the higher \ac{DoX}, the greater the chances to adequately explain to a variety of users.
	
	Secondly, the numerical differences between the \ac{DoX} scores shown in table \ref{tab:exp1_dox} and \ref{tab:exp2_dox} suggest that our algorithm for computing \ac{DoX} scores may be sensitive to the choice of a deep language model for pertinence estimation.
	In fact, on the one hand, we see that the difference in terms of \ac{DoX} between the normal \textit{XAI-based explainers} and the other explainer tend to differ from MiniLM to the Universal Sentence Encoder slightly.
	Nonetheless, we also see that in all the considered experiments, the \ac{DoX} scores increase as expected, with both MiniLM and the Universal Sentence Encoder, suggesting that the alignment of \ac{DoX} to \textit{explainability} is independent of the chosen deep language model.
	This intuition is supported by the fact that the deep language models, on average, perform reasonably well on existing benchmarks for evaluating answer retrieval algorithms.
	In other words, if the average \ac{DoX} aggregates enough archetypes, aspects, and details, then different pertinence functions performing similarly on standard benchmarks may produce proportionally similar scores.
	This does not exclude the fact that some deep language models might be better than others for computing \ac{DoX} scores or that multiple standardized deep language models should be adopted for a thorough estimate of the \ac{DoX}. 
	We leave this analysis for future work.
	
	Another possible limitation of \ac{DoX} is that its scores cannot be easily normalized in a $[0,1]$ range.
	In fact, according to Definition \ref{def:DoX}, \ac{DoX} is computed by performing a sum (called \textit{cumulative pertinence}) over the set of details $D$ extracted from an \textit{explanandum support material}, so that \ac{DoX} can measure the similarity of the \textit{explanandum support material} to the explanandum.
	Unfortunately, it is impossible to know the total number of details of any possible \textit{explanandum support material}. Therefore, it is impossible to normalize the score by dividing the \textit{cumulative pertinence} by such number. 
	It is worth noting that such a sum is necessary. 
	Indeed, suppose the \textit{cumulative pertinence} were a mean instead of a sum. In that case, the resulting score for an \textit{explanandum support material} could not be compared to that of any larger (in terms of the number of details) \textit{explanandum support material}, making pointless the use of \ac{DoX} in the first place.
	
	Furthermore, it is essential to mention that \ac{DoX}, alone, is not sufficient for a thorough quantification of how much of the information is explained by an AI.
	Our definition of \ac{DoX} does not consider the correctness of information of the \textit{explanandum support material}, assuming that truth is given and that it is different from explainability.
	In other words, \ac{DoX} should always be used with other metrics that can evaluate the correctness of available information.
	
	Finally, although \ac{DoX} can be used to verify many of the requirements defined by \cite{bibal2021legal}, it is still unclear how to apply \ac{DoX} to verify also Government-to-Citizen legal requirements. 
	Selecting a reasonable threshold of \ac{DoX} scores for law compliance is undoubtedly one of the challenges we envisage for a proper standardization of \textit{explainability} in the industrial context. 
	We also leave these analyses for future work.
	
	\section{Conclusions} \label{sec:conclusions}
	
	In this paper, we proposed a new metric for explainability called \ac{DoX} that could objectively quantify how much of the information is explained by an AI.
	For instance, \ac{DoX} can be used to verify the satisfaction of Business-to-Business, and Business-to-Consumer requirements as defined by \citeauthor{bibal2021legal} \cite{bibal2021legal}.
	
	\ac{DoX} is based on the intuition coming from Achinstein's theory of explanations that explaining is an act of illocutionary question-answering.
	Specifically, \ac{DoX} frames explanations as answers to many simple questions (\textit{archetypes}), shedding light on the concepts being explained so that the more (archetypal) answers a corpus can give about essential aspects of an explanandum, the more that corpus is explainable.	
	Thus \ac{DoX} is the first explainability metric based on Ordinary Language Philosophy. It is a model-agnostic and deterministic approach that can work with any corpus of explainable information represented in natural language (i.e., English).
	
	In particular, \ac{DoX} quantifies the three main criteria of explainability adequacy defined by Carnap: similarity, exactness, and fruitfulness.
	In this sense, our contribution is a mechanism for quantifying Carnap's criteria and aggregating them together in one single score called average \ac{DoX}, used to compare the degree of explainability of different explanatory systems.
	\ac{DoX} can quantify the degree of explainability of a corpus of information by estimating how adequately that corpus could answer an arbitrary set of archetypal questions about the concepts of an explanandum.
	
	Throughout the paper, we also presented a concrete implementation of \ac{DoX} called DoXpy.
	
	
	In order to understand whether the \ac{DoX} is behaving as expected, we designed a few experiments on two realistic systems for heart disease prediction and credit approval, involving state-of-the-art AI technologies such as Artificial Neural Networks, TreeSHAP \cite{lundberg2020local}, XGBoost \cite{chen2016xgboost}, and CEM \cite{dhurandhar2018explanations}.
	The results show that the \ac{DoX} is aligned with our expectations and can be used to quantify \textit{explainability} in natural language information corpora.
	
	Although \ac{DoX} cannot be used directly on a black-box model to understand how much of it can be explained, it can be used on the output of an ensemble of \ac{XAI} algorithms or any other explainable information (e.g., documentation, papers, books) to understand how that information can be used to explain.
	In this sense, \ac{DoX} is the most useful when used to evaluate extensive collections of explainable information (e.g., the output of an ensemble of \ac{XAI} algorithms).
	
	Another context for applying \ac{DoX} could be education. Not surprisingly, many would argue that explanations are one of the primary artifacts through which humans understand reality and learn to solve complex problems \cite{berland2009making}.
	Therefore, \textit{explaining} is central to \ac{XAI} and education, and these are two contexts where our technology and understanding of explanations could be of utmost importance.
	
	\bibliographystyle{cas-model2-names}
	\bibliography{biblio}
	
	\begin{acronym}
		\acro{XAI}{Explainable AI}
		\acro{DoX}{Degree of Explainability}
	\end{acronym}
	
\end{document}